\PassOptionsToPackage{capitalize}{cleveref}

\documentclass[a4paper]{article}

\usepackage[utf8]{inputenc} \usepackage[T1]{fontenc}    \usepackage{hyperref}       \hypersetup{
        colorlinks   = true,
        urlcolor     = blue,
        linkcolor    = blue,
        citecolor   = blue
}
\usepackage{url}            \usepackage{booktabs}       \usepackage{amsfonts}       \usepackage{nicefrac}       \usepackage{microtype}      \usepackage{xcolor}         

\usepackage{changepage}
\usepackage{wrapfig}
\usepackage{microtype}
\usepackage{graphicx}
\usepackage{subcaption}
\usepackage{booktabs} 

\usepackage{xcolor}
\usepackage[boxed]{algorithm}
\usepackage{algorithmic}
\usepackage{amsmath}
\usepackage{amsthm}
\usepackage{amssymb}
\usepackage{bbm}
\usepackage{comment}
\usepackage{authblk}
\usepackage{placeins}
\usepackage{caption}
\usepackage{multirow}
\usepackage{xspace}
\usepackage{spverbatim}
\usepackage{csquotes}
\usepackage{float}
\usepackage{listings}

\usepackage{mathrsfs}

\usepackage[capitalize]{cleveref}
\usepackage[nolist,nohyperlinks]{acronym}

\usepackage{dsfont}

\usepackage{enumitem}

\usepackage{mathtools}

\usepackage{array}
\usepackage{tabularx}

\usepackage{fullpage}

\usepackage{natbib}
\setcitestyle{round,sectionbib}

\newcommand{\pr}[1]{\left( #1 \right)}

\newcommand{\cbr}[1]{\left\{ #1 \right\}}
\newcommand{\abs}[1]{\left|#1\right|}

\newcommand{\tp}{^{\top}}
\newcommand{\ip}[1]{\left\langle #1 \right\rangle}

\newcommand{\ind}{\mathbf{1}}

\newcommand*\diff{\mathop{}\!\mathrm{d}}

\DeclareMathOperator*{\argmin}{arg\,min}

\DeclareMathOperator*{\diag}{diag}
\DeclareMathOperator*{\tr}{tr}

\newcommand{\vertiii}[1]{{\left\vert\kern-0.25ex\left\vert\kern-0.25ex\left\vert #1 
    \right\vert\kern-0.25ex\right\vert\kern-0.25ex\right\vert}}

\newcommand{\R}{\mathbb{R}}
\newcommand{\ve}{\varepsilon}

\newcommand{\cN}{\mathcal{N}}

\renewcommand{\th}{\theta}

\newcommand{\sumin}{\sum_{i=1}^n}

\renewcommand{\vec}{\mathrm{vec}}
\renewcommand{\d}{\diff}
 \begin{acronym}
\acro{RLS}{Regularized Least Squares}
\acro{ERM}{Empirical Risk Minimization}
\acro{RKHS}{Reproducing kernel Hilbert space}
\acro{PD}{Positive Definite}
\acro{PSD}{Positive Semi-Definite}
\acro{SGD}{Stochastic Gradient Descent}
\acro{OGD}{Online Gradient Descent}
\acro{GD}{Gradient Descent}
\acro{SGLD}{Stochastic Gradient Langevin Dynamics}
\acro{IS}{Importance Sampling}
\acro{MGF}{Moment-Generating Function}
\acro{ES}{Efron-Stein}
\acro{ESS}{Effective Sample Size}
\acro{KL}{Kullback-Liebler}
\acro{SVD}{Singular Value Decomposition}
\acro{PL}{Polyak-Łojasiewicz}
\acro{NTK}{Neural Tangent Kernel}
\acro{NTRF}{Neural Tangent Random Feature}
\acro{ReLU}{Rectified Linear Unit}
\acro{ES}{Efron-Stein}
\acro{whp}[w.h.p.]{with high probability}
\acro{wrt}[w.r.t.]{with respect to}
\acro{wlog}[w.l.o.g.]{without loss of generality}
\acro{ResNet}{Residual Network}
\acro{NTF}{Neural Tangent Feature}
\acro{GF}{Gradient Flow}
\end{acronym}
 
\newtheorem{theorem}{Theorem}

\newtheorem{lemma}{Lemma}
\newtheorem{corollary}{Corollary}
\newtheorem{remark}{Remark}
\newtheorem{proposition}{Proposition}

\usepackage{times}

\begin{document}

\title{Low-rank bias, weight decay, and model merging in neural networks}

\author[1]{Ilja Kuzborskij}
\author[2]{Yasin Abbasi Yadkori}
\affil[1]{Google DeepMind}
\affil[2]{Sapient Intelligence}
\date{} 
\setcounter{Maxaffil}{0}
\renewcommand\Affilfont{\itshape\small}

\maketitle

\begin{abstract}
  We explore the low-rank structure of the weight matrices in neural networks
  at the stationary points (limiting solutions of optimization algorithms)
with $L2$ regularization (also known as weight decay). We show several
  properties of such deep neural networks, induced by $L2$ regularization.
  In particular, for a stationary point we show alignment of the
  parameters and the gradient, norm preservation across layers, and low-rank
  bias: properties previously known in the context of solution of gradient descent/flow type algorithms.  Experiments
  show that the assumptions made in the analysis only mildly affect the
  observations.

  In addition, we investigate a multitask learning phenomenon enabled by $L2$
  regularization and low-rank bias. In particular, we show that if two networks
  are trained, such that the inputs in the training set of one network are
  approximately orthogonal to the inputs in the training set of the other
  network, the new network obtained by simply summing the weights of the two
  networks will perform as well on both training sets as the respective
  individual networks.  We demonstrate this for shallow ReLU neural networks
  trained by gradient descent, as well as deep linear networks trained by gradient flow.
\end{abstract}

\section{Introduction}
\label{sec:intro}

First-order optimization algorithms, such as \ac{SGD} have emerged as a go-to
tool for training machine learning models.  Their popularity primarily stems
from good scalability and empirical performance, while their theoretical
properties and performance are reasonably well understood for learning problems
with sufficient structure, such as convexity or even weak forms of
non-convexity.  However, in the recent years with the advent of
overparameterized deep neural networks, there was a surge of interest in
understanding the behavior of these algorithms on non-convex and often
non-differentiable problems. Here, a combination of neural network architecture
choice, initialization, and hyperparameter tuning often achieves near-zero
training loss while enabling good test-time generalization ability.

Recently, some progress was made in attempt to explain this phenomenon by
looking for \emph{implicit biases} in the algorithm~\citep{vardi2023implicit}.
While implicit biases were known before in the context of simpler learning
problems such as overparameterized linear least-squares (solving this problem
through pseudo-inverse gives an interpolating solution
with the
\emph{minimal $L2$ norm}), observing the same phenomenon in neural network
learning is rather recent.  \citet{lyugradient,ji2020directional} showed that for
a certain family of neural networks (\emph{positive homogeneous} neural
networks, such as deep linear or deep ReLU neural networks) trained by an idealized version of \ac{GD} known as \ac{GF},
asymptotically converges to the predictor with the smallest parameter $L2$ norm.
In another influential line of work, known as the \ac{NTK}
approximation~\cite{jacot2018neural,du2018gradient,ji2019polylogarithmic},
it was shown that a sufficiently wide shallow non-linear neural network behaves
similarly as a linear predictor on \ac{RKHS}.  This enabled reduction to
analysis of \ac{GD} on \ac{RKHS}, which is known to converge to the minimum $L2$
norm interpolating solution (or, which behaves as a regularized solution when
stopped early~\citep{yao2007early}).  

While $L2$ norm minimization (or regularization) has a clear interpretation in
linear models, its role in deep neural network learning is less intuitive.  On
that front, several works showed that this form of regularization induces a
\emph{low-rank} structure in weight matrices of neural networks.
In fact, in deep linear neural networks weight matrices in such interpolants are
shown to be rank-$1$ matrices~\citep{ji2019gradient}.  Similarly,
\citet{timor2023implicit} showed that weight matrices in minimum $L2$ norm
interpolating deep ReLU networks will have a low \emph{stable rank} (ratio
between Frobenius and spectral norms) as long as the depth is sufficiently
large. See \Cref{sec:related} for other related works.

Finally, while the appearance of low-rank weight matrix structure during
training is interesting on its own, here we might ask a related question whether
low-rank bias can explain other phenomena that we observe in neural network
learning.  In particular, in this paper we look at the connection between
low-rank bias and a form of a multi-task learning known as \emph{model merging}.
Here, weight matrices of two neural networks trained on different
tasks are summed to form a new neural network, which often performs well on both
original tasks.  This behaviour seems surprising at first glance, and the
reason why model merging (or, related, parameter averaging)
might
be effective is not well understood.

\subsection{Our contributions}
In this paper we take a closer look at a low-rank bias and its effect on model merging.
Our contribution is two-fold yet at the core it is connected to low-rank bias induced
by $L2$ regularization. \paragraph{Low-rank bias and alignment through weight decay}
Model merging discussed earlier crucially relies on $L2$ regularization, and its success (good performance on both tasks) can be attributed to biases that arise in neural networks because of regularization. To this end, in \Cref{sec:stationary} we take a closer look at such biases.
In particular, we focus on depth-$K$ neural networks with $H$-positive homogeneous activations (such as ReLU or powers of ReLU)
and consider stationary points of the empirical logistic loss with $L2$ regularization.
In this context, stationary points enjoy \emph{alignment}, meaning that
parameters converge in the direction of gradient of the
(unregularized) loss.
Asymptotic alignment was known before in case of
\ac{GF} and \ac{GD} (without regularization)~\citep{ji2019gradient,ji2020directional}.
Building upon this observation we show
the following results:
\begin{itemize}
\item Near norm-preservation: weight matrices in neural networks we consider have the same Frobenius norm throughout layers, up to a scaling factor governed by the homogeneity parameter $H$ and layer index.
\item Deep linear neural networks have all weight matrices of rank one.
\item Deep (possibly non-linear) neural networks have a low-rank weight matrix
  structure, in the following sense:
\end{itemize}
The weighted harmonic mean of stable ranks, or \emph{pseudo-rank}\footnote{A stable rank of matrix is defined as
  a ratio of its Frobenius and spectral norms. It is small whenever matrix
  has few dominant eigenvectors.} of weight matrices $(W_1, \ldots, W_K)$ is
controlled by regularization parameter $\lambda$ and the training loss at the non-zero stationary point:
\begin{align*}
  \text{Pseudo-rank} =
  \frac{1}{\text{weight}_1 \left(\frac{\| W_1\|_F}{\|W_1 \|_2}\right)^{-1} + \dots + \text{weight}_K \left(\frac{\| W_K\|_F}{\|W_K \|_2}\right)^{-1}}
  \leq
  \sqrt{\frac{\text{(Training loss)}^{\alpha}}{\lambda}}
\end{align*}
for some $\alpha > 1/2$ that depends only on $(H, K)$, and
where weights $(\text{weight}_k)_{k=1}^K$, depending only on $(H, [K])$, increase monotonically as the layer index $k$ increases, entails that we effectively control the harmonic mean of the last layers.

The bound on the pseudo-rank is problem-dependent in a sense that it scales with the training loss.
Therefore considering different solutions (stationary points of the regularized loss) we can observe different behaviours of the pseudo-rank. 
For instance, assume that the regularized loss at the stationary point is no greater than the regularized loss at the $0$ vector, which is for instance satisfied by the global minimizer.
Then, for any $\lambda$ that ensures non-zero global minimizer,
\begin{align*}
  \text{Pseudo-rank} \leq \sqrt{1 / \lambda}
\end{align*}
that is pseudo-rank decreases with increase of $\lambda$.
Another example, is to consider the limiting training error of some optimization algorithm with step size appropriate tuning, and Xavier-type initialization.
In this case, 
\begin{align*}
  \text{Pseudo-rank} \lesssim \sqrt{1 / \lambda + \text{(depth)} \cdot \text{(width)}}~.
\end{align*}

Harmonic mean of stable ranks as a proxy for capturing the rank structure in
deep neural networks was proposed by \cite{timor2023implicit}. Since stable rank
cannot be smaller than one, the harmonic mean is small whenever majority of
weight matrices have small stable ranks.  In particular,
\cite{timor2023implicit} looked at the minimum $L2$-norm \emph{interpolating}
neural networks, in a sense that for all training examples $y_i \in \{\pm 1\}$,
$y_i \cdot \mathrm{prediction}(x_i) \geq 1$.
Their conclusion was that the harmonic mean $\to \sqrt{2}$ as depth
$K \to \infty$.  In contrast, we simply require \ac{GD} to achieve a stationary
point, in which case the harmonic mean is controlled by regularization parameter
and the training loss at stationarity, even at a finite depth (see
\Cref{lem:low-rank} and discussion therein).  Indeed, this seems to be supported
by some basic empirical evidence (see \Cref{sec:norm-rank-exps,fig:rank-l2}),
which suggests that that a low stable rank can be achieved even without
interpolation but with weight decay.

Finally, we look at the behavior of a stable rank and norm preservation
empirically for very deep neural networks (such as pre-trained large language
models) in \Cref{sec:norm-rank-exps}, and conclude that assumptions (such as
homogeneity and stationarity) only mildly affect the observations.
\paragraph{Model merging enabled by weight decay.} Next, we show that low-rank bias in
neural networks enables effective \emph{model merging}: after neural networks
are trained on different datasets with mutually nearly-orthogonal inputs (but
not necessarily orthogonal within the task), simply summing their weight
matrices results in a predictor with combined weights that performs well
simultaneously on \emph{all tasks}.  We explain this phenomenon through
low-rank bias.
Training biases different networks toward low-rank weight matrices that span
non-overlapping subspaces.  Consequently, summing their weight matrices results
in a neural network that behaves as the original ones on inputs from respective
tasks.
In other words, two neural networks can be made to `reside' in one parameter
set.

More formally, given input-label pairs
$(x_1, y_1), \ldots, (x_n, y_n) \in \mathbb{S}^{d-1} \times \{\pm 1\}$ we obtain
a predictor $f_{\th(t)}$, whose parameters $\th(t)$ are found by \ac{GD} (or
\ac{GF}) run for $t$ steps, while minimizing regularized loss
$\th \mapsto \frac1n \sum_i \mathrm{loss}(f_{\th}(x_i), y_i) + \lambda \|\th\|^2$.
In a similar way, we obtain parameters $\th'(t)$ given another data
$(x_1', y_1'), \ldots, (x_n', y_n')$ belonging to the second task, such that
inputs between these tasks are approximately orthogonal
in a sense that
$\max_{i,j} \abs{\ip{x_i, x_j'}} \leq \ve$.  Then, given
an input $x$ originating from the first task, we show that
\begin{align*}
  |f_{\th(t)}(x) - f_{\th(t) + \th'(t)}(x)| \leq f_{\th'(0)}(x) \, e^{-\lambda t} + C_2 \, \ve
\end{align*}
for several scenarios.
Here, exponentially vanishing term
that appears because of $L2$ regularization, is responsible for contribution of
initialization, which is typically non-zero in neural network training.  The
second, $\ve$-dependent term captures the length of projection of the input (or
activation vector in case of multilayer neural networks) from one task onto the
weight matrix of the network trained on another task.  Note that $L2$
regularization is essential here, since without it, the effect of initialization
(appearing through a constant term $f_{\th'(0)}(x)$) would not disappear.
Interestingly, none of these results require convergence to the local minimum.

In particular, we show the above in case of linear prediction and shallow ReLU
neural networks trained by \ac{GD}, and deep linear neural networks trained by
\ac{GF}.
Finally, in \Cref{sec:experiments:merging} we experimentally show
that the same findings hold in case
of fully-connected ReLU neural networks, which suggests that the same conclusions hold beyond our theoretical results.
\section{Definitions and preliminaries}
\label{sec:def}
Throughout, $\|\cdot\|$ is understood as the Euclidean norm for vectors and the Frobenius norm matrices.
When, written explicitly for matrices, $\|\cdot\|_2$ is a spectral norm, and $\|\cdot\|_F$ is a Frobenius norm.
For some matrix $A$ its \emph{stable rank} is defined as a ratio $\|A\|_F / \|A\|_2$.
The Frobenius inner product between matrices $A,B$ is denoted by
$\ip{A, B} = \tr(A\tp B)$.
Throughout $e_j = (\ind(j=1), \ind(j=2), \ldots, \ind(j=m)) \in \{0,1\}^m$
and $a \wedge b = \min(a, b)$ and $a \vee b = \max(a,b)$.

In the following we denote \ac{ReLU} operation by $(x)_+ = \max(x, 0)$ for
$x \in \R$.  For vectors and matrices application of $(\cdot)_+$ is understood
elementwise.  \ac{ReLU} has several useful properties.  For $a,b \in \R$, we
have $|(a)_+ - (b)_+| \leq |a-b|$ and so for vectors $x,y \in \R^d$,
$\|(x)_+ - (y)_+\|_2^2 = \sum_i ((x_i)_+ - (y_i)_+)^2 \leq \sum_i (x_i - y_i)^2
= \|x - y\|_2^2$.

Function $f : \R^d \to \R$ is called \emph{positive homogeneous} of degree $K$ when $f(\alpha x) = \alpha^K f(x)$ for any $\alpha \geq 0$.
Euler's homogeneous function theorem states that, if the chain rule holds, then $K f(x) = \ip{x, \nabla f(x)}$.
Note that \ac{ReLU} is positive homogeneous, meaning that $(\alpha x)_+ = \alpha (x)_+$ for $\alpha \geq 0$.
In particular, this implies that for a $K$-layered \ac{ReLU} or linear neural network is positive homogeneous and so $f_{\th}(\alpha \, x) = \alpha^K f_{\th}(x)$ and so Euler's theorem holds.

For the logistic loss function $\ell(x) = \ln(1+e^{-x})$ we have
$\ell'(x) = - e^{-x} / (1 + e^{-x}) \in (-1, 0)$, and moreover the following facts hold:
$|\ell'(x)| = -\ell'(x) \leq \ell(x)$,
$x \, \ell'(x) \leq \ell(x)$,
and $-x \, \ell'(x) \leq \sqrt{\ell(x)}$
for all $x$.

\paragraph{Differentiation}
We introduce some formalism for non-differentiable functions since occasionally
we will work with the \ac{ReLU} activation.  For some $F : \R^p \to \R$ that is locally Lipschitz\footnote{$F$ is locally Lipschitz if for every point $\th$, there exists a neighborhood $B \supseteq \{\th\}$ such that $F$ is Lipschitz on $B$.}, we denote by $\partial F$ its Clarke
differential:
\begin{align*}
  \partial F(\th) = \mathrm{conv}\pr{\cbr{
    g \in \R^p : \exists (\th_i)_i \to \th, \quad \nabla F(\th_i) \to g
  }}.
\end{align*}
Vectors in $\partial F$ are called subgradients, while $\bar \partial f(x)$ will
stand for a unique minimum-norm subgradient, that is
$ \bar\partial f(x) = \argmin_{g \in \partial f(x)} \|g\|~ $.  Throughout the
paper it is understood that $\nabla f = \bar\partial f$.
We will assume that the chain rule holds, that is
$\dot f(\th(t)) = \langle g(t), \dot \th(t) \rangle,$ for all
$g(t) \in \partial F(\th(t))$.  It is possible to formally establish that the
chain rule holds by assuming a technical notion of `definability' (not covered
here, see for instance \citep{ji2019gradient}), which excludes functions that might result in
a badly behaved optimization.

\paragraph{Neural networks}
\label{sec:reliminaries}

In this paper we consider multilayer neural networks $f_{\th} : \R^d \to \R$ given by a recursive relationship
\begin{align*}
  f_{\th}(x) = \ip{w_K, h_{K-1}}, \quad h_k = \sigma(W_k h_{k-1}), \quad h_0 = x \qquad (k \in [K-1])
\end{align*}
where $x$ is the input, $W_1, \ldots, W_K \in \R^{m \times m}$
is collection of weight matrices, $\th = (\vec(W_1), \ldots, \vec(W_K))$ is a parameter vector, and $\sigma : \R^m \to \R^m$ is activation function.

In practice parameters $\th$ are tuned based on the training data by minimizing some loss function.
In this paper we will focus on a binary classification problem, and so given inputs and labels $(x_i, y_i)_{i=1}^n \in (\mathbb{B}^d \times \{-1,1\})^n$, the training loss and its regularized version are defined as
\begin{align*}
  L(\th) = \frac1n \sumin \ell\pr{y_i \, f_{\th}(x_i) }~, \qquad L_{\lambda}(\th) = L(\th) + \frac{\lambda}{2} \|\th\|_2^2 \qquad (\lambda \geq 0)
\end{align*}
where $\ell()$ is a logistic loss function $\ell(x) = \ln(1+e^{-x})$.
In this work we consider standard
\emph{\ac{GD}} algorithm which is often used to approximately minimize $L_{\lambda}$ through
recursive updates, for all $k$,
\begin{align*}
      W_{k,s+1} = (1-\eta \lambda) W_{k,s}
  - \frac{\eta}{n} \sumin y_i \, \ell'(y_i \, f_{\th_s}(x_i)) \, \frac{\diff f_\th(x_i)}{\diff W_k}(\th_s)
  \qquad (s=0,1,2,\ldots,t)
\end{align*}
where $\eta > 0$ is a step size and $\th_0$ is an initial parameter vector.

When we consider \ac{GF} dynamics (we use $(t)$ time indexing instead of $\cdot_t$ as in the discrete case), the update rule is replaced by time derivative
\begin{align}
\label{eq:GF-W}
  \dot W_k(t) = -\lambda W_k(t) - \frac1n \sumin y_i \, \ell'(y_i \, f_{\th(t)}(x_i)) \, \frac{\diff f_\th(x)}{\diff W_k}(\th(t))~.
\end{align}

\section{Stationary points and low-rank bias}
\label{sec:stationary}
In this section we consider training deep neural networks with positive homogeneous activation functions (such as identity or ReLU or smoothed versions of ReLU, e.g.\ powers of ReLU), with weight decay regularization.
Proofs for all statements in this section can be found in \Cref{proof:sec:stationary}.
In particular, we study solutions $\theta$ that are the \emph{stationary points} of the regularization empirical loss $L_{\lambda}$.
These solutions satisfy the following \emph{alignment condition}:\footnote{When $L_{\lambda}$ is non-smooth (for example when activation function is ReLU), $\nabla$ is understood the Clarke subgradient. In this case, formally
  $\lambda \theta = - g$ for $g \in \partial L_{\lambda}(\theta)$.
}
\begin{align}
  \label{eq:alignment}
  \lambda \theta = - \nabla L(\theta)~.
\end{align}
Under standard smoothness and boundedness conditions, first-order methods such as SGD, Adam, and AdaGrad produce iterates whose gradients of the regularized objective $L_{\lambda}$ vanish in expectation/asymptotically~\citep{defossez2022simple,li2023convergence}; hence alignment holds at their limit points. For non-smooth objectives (e.g., ReLU nets), these guarantees do not apply in full generality; one typically works with Clarke stationarity or adds additional structure/weak convexity.
For the non-smooth $L_{\lambda}$ there exist hardness results for deterministic algorithms to efficiently achieve $(\delta, \varepsilon)$-stationary points~\citep{kornowski2021oracle}.
These worst-case lower bounds apply to deterministic black-box methods and do not preclude the behavior observed with stochastic optimizers used in practice.
However, here we treat \cref{eq:alignment} as an idealization when taking $\ve \to 0, \delta \to 0$.

This type of alignment was first studied by \citep{ji2019gradient,ji2020directional} in the context of training with gradient flow, and under the extra condition that $L(\theta_0) < \ell(0)$. (They also show alignment for GD for deep linear networks using an adaptive step-size and under the extra conditions that the initialization has sufficiently small loss.) Here, weight decay enables a simpler argument to establish alignment.
Next, we show several implications of this result.
\begin{lemma}[Deep linear networks]
\label{lem:deep-linear}
Let $f_{\theta}$ be a deep linear network, and $\theta$ be a parameter vector satisfying the alignment condition of~\cref{eq:alignment}. Then all weight matrices $W_1,\dots,W_{K-1}$ are rank-1.
\end{lemma}
\begin{lemma}[Norm preservation]
\label{lem:norm-preservation}
Let $\theta$ be a parameter vector satisfying the alignment condition of~\cref{eq:alignment}. Further, assume that $f_\theta$ is locally Lipschitz and with $H$-positively homogeneous activations. Then weight matrices have Frobenius norm:
$$
\lambda \|W_k \|_F^2 = - \frac{H^{K-k}}{n} \sum_i  \ell'(y_i f_{\theta}(x_i)) y_i f_{\theta}(x_i) \;.
$$
\end{lemma}
A deep neural network with non-linear activations can satisfy the above conditions, for instance in case of ReLU $H=1$.
\begin{lemma}[Low-rank structure]
  \label{lem:low-rank}  
  Consider  logistic ($\ell(x)=\ln(1+e^{-x})$) loss function. Assume activations are $H$-positive homegeneous and satisfy $\|\sigma(v)\| \leq \|v\|^H$. Assume that $K \geq 2, H \geq 1$.
  For any $\lambda > 0$ such that $\theta$ is a non-zero stationary point of $L_{\lambda}$ we have that
\[
  \sum_k \frac{(H^{K-k})^{3/2}}{Z} \cdot \frac{\|W_k\|_2}{\|W_k\|_F} \geq \sqrt{\lambda} \, L(\theta)^{-(\frac14 + \frac{1}{2 Z})}
\]
where $Z = \sum_{k} H^{K-k}$. Note that $Z = K$ for $H=1$ and $Z = (H^K-1)/(H-1)$ otherwise.
\end{lemma}
The above result implies that the average stable rank decreases as $\lambda$ increases.
In particular, to show that the lower bound increases with $\lambda$ it is enough to upper bound $L(\theta)$.
\begin{corollary}
  Assume that $L_{\lambda}(\theta) \leq L_{\lambda}(\theta_0)$ for some fixed $\th_0$ and that $L(\theta_0) \geq 1$.  
  Then, under conditions of \Cref{lem:low-rank}\footnote{Note that the lower bound holds since $L(\theta_0) \geq 1$ by assumption and
since for $H \geq 1, K \geq 2$, we have
$1/Z = (H-1)/(H^K-1) \leq 1/K \leq 1/2$.}
  $$
  \sum_k \frac{(H^{K-k})^{3/2}}{Z} \cdot \frac{\|W_k\|_2}{\|W_k\|_F}
\geq
\sqrt{\frac{\lambda}{L(\theta_0) + \lambda \|\theta_0\|^2}}~.
$$
\end{corollary}
For instance, when $\th$ is a global minimizer of $L_{\lambda}$, it is easy to verify that $L(\theta) \leq L_{\lambda}(\theta) \leq L_{\lambda}(0) \leq 1$, and so the lower bound implied by \Cref{lem:low-rank} becomes $\sqrt{\lambda}$ for $\lambda \leq C_{H,K}$.\footnote{Where $C_{H,K} = (\sum_k(H^{K-k})^{3/2} / Z)^2$.}

Another very basic example is when $\theta_0$ is initialization of some optimization algorithm which does not increase the objective (e.g. such as \ac{GD}).
In this case, the lower bound depends on width and depth, e.g. $\|\th_0\|^2 \sim K m$ for Xavier initialization.

\begin{remark}
  LHS in \Cref{lem:low-rank} is bounded by a constant whereas RHS is $\lambda$-dependent.
  One can imagine that by taking arbitrarily large $\lambda$ on the RHS we could achieve inconsistency.
  This cannot happen because permissible range of $\lambda$ is restricted by a `non-zero $\theta$' condition.
  Indeed, the following sketch argues that the unique global solution of $L_{\lambda}$ for large enough $\lambda$ must be $\theta = 0$.

For simplicity consider the case $H=1$ (ReLU or linear activation).
By AM-GM inequality $|f_{\th}(x)| \leq \|\th\|^K K^{-K/2}$ (see the first step of the proof of \Cref{lem:low-rank}) and
consider the proxy to $L_{\lambda}$, which employs the upper bound on $|f_{\th}(x)|$:
\begin{align*}
  g_{\pm}(\|\th\|) = \frac{\lambda}{2} \|\th\|^2 + \ln(1 + \exp(\pm \|\th\|^K K^{-K/2}))
\end{align*}
which `sandwiches' $L_{\lambda}$.
It is not hard to check that for $K \geq 2$, and $\lambda > C_{H,K}$ (constant that depends only on $K$ and $H$), $g'(x) = 0$ has one solution which is $0$.
Note that this is specific for multilayer neural networks as this phenomenon does not occur with $K=1$

The intuition is that the power $K$ in $\|\th\|^K$ makes the loss function concentrate around zero (because $\cdot^K$ makes predictions larger and logistic loss becomes tiny), but in the vicinity of zero $\ell_2$ penalty is stronger and the only surviving global minimum is the one of the $\ell_2$ penalty.
\end{remark}

\citet{timor2023implicit} show low-rank structures for the global optimum of the minimum-rank interpolating solution assuming existence of a smaller ``teacher'' network with $K'<K$ layers that interpolates data and Frobenius norm of its weight matrices are bounded by $C$. More specifically, they show that $(1/K) \sum_k \| W_k\|_2 / \|W_k \|_F \ge \left(1/C\right)^{K' / K}$. We can have an intuitive understanding of their result by noting that given a smaller interpolating network, the last layer of the network tends to be low-rank due to the \textit{Neural Collapse} phenomena~\citep{papyan2020prevalence}. Then if we add more layers, the additional layers will remain low-rank as their activations will lie in a low-rank structure, and $(1/K) \sum_k \| W_k\|_2 / \|W_k \|_F$ will approach one as we add more layers. 

Results in \Cref{lem:deep-linear} and \Cref{lem:low-rank} show a different phenomena. Remarkably, in both these results, irrespective of network capacity or performance, and as a consequence of weight decay, the weight matrices will become low-rank on average. Another interesting feature is that, unlike results of \citet{timor2023implicit}, depth plays a minor role in establishing low-rank structures in \Cref{lem:deep-linear} and \Cref{lem:low-rank}. In \Cref{sec:norm-rank-exps}, we present experiments that show that even if number of layers is small and network has a large number of classification mistakes, the average inverse stable-rank grows as weight decay parameter $\lambda$ increases.   

Finally, low-rank results we showed here also to some extent apply to modern architectures such as attention heads, which are fundamental building blocks of transformers~\citep{vaswaniattention}.
In particular, if we replace $\mathrm{softmax}$ with $\mathrm{argmax}$, and remove residual connections and layer normalization, we will obtain a \emph{homogeneous attention head}, and \Cref{lem:norm-preservation,lem:low-rank} hold for such models. Transformers used in practice typically include residual connections, layer norms, softmax attention, and so on, that deviate from the above conditions. 
Yet, as we show empirically in \Cref{sec:norm-rank-exps}, the conclusion of the above result holds to some extent.

\section{Merging model parameters}
\label{sec:merging}
Throughout this section, in addition to the training tuple $(x_i, y_i)_{i=1}^n$ we introduce $(x_i', y_i')_{i=1}^n$, and in addition to the regularized objective $L_{\lambda}$ we will introduce $L_{\lambda}'$ defined with respect to the second training set.
As an warm-up example we first consider a linear prediction scenario where we
are looking at predictors of a form $f_\th(x) = \ip{\th, x}$ and suppose that
$\th_t$ is obtained by \ac{GD} minimizing $L_{\lambda}$, while
$\th_t'$ is obtained by minimizing $L_{\lambda}'$.  Now given a
test point $x$ that is sufficiently different from inputs $(x_i')_{i=1}^n$, or assuming that $\max_i |\ip{{x_i'}, x}| \leq \ve$ we have
\begin{align*}
  f_{\th_t + \th_t'}(x) \approx f_{\th_t}(x)
\end{align*}
and the same holds for some point $x'$ sufficiently different from $(x_i)_{i=1}^n$.
This is derived in a straightforward way based on \ac{GD} update rule
$
  \th_{t+1} = (1- \lambda \eta) \th_t - (\eta/n) \sumin \ell'(y_i \ip{\th_t, x_i}) x_i y_i
$
while unrolling the recursive relationship we get
\begin{align*}
  \th_{t}
   &= \th_0 (1- \eta \lambda)^t + \sumin \alpha_i x_i \quad
  \text{where} \quad \alpha_i = \frac1n \sum_{s=0}^{t-1} \eta (1-\eta \lambda)^{t-s-1} \ell'(y_i \ip{\th_s, x_i}) y_i~.
\end{align*}
The above identity tells us that the solution
$\th_t$ resides in the span of inputs.
So, it is easy to see that prediction on the inputs from the other task is close to $\ve$
  \begin{align*}
    \abs{\ip{\th_t, x'}}
    &\leq
      \abs{\ip{\th_0, x'}} (1- \eta \lambda)^t + \ve \, \frac{1 - (1-\eta \lambda)^t}{\lambda}
  \end{align*}
  for $1$-Lipschitz loss function.
In particular, the above implies that the gap of interest $|f_{\th_t + \th_t'}(x) - f_{\th_t}(x)|$ is also controlled by the upper bound in the display above.
  Note that a possibly large term $\abs{\ip{\th_0, x'}}$ is attenuated
  exponentially quickly by the weight decay, so its effect is negligible at the
  end of optimization.  While in the linear case we could set $\th_0 = 0$, in
  case of neural network learning setting initialization at $0$ is
  atypical and therefore weight decay seems to have an important role in such
  scenarios.
  Another summand on the right hand side is $\ve$-dependent and captures similarity between inputs from different tasks. If inputs are orthogonal this term disappears.
  
\paragraph{Shallow \ac{ReLU} networks}
Here we consider a shallow \ac{ReLU} neural network
\begin{align}
  \label{eq:shallow}
  f_\th(x) = \ip{u, (W x)_+} \qquad (x \in \R^d)
\end{align}
where hidden weight matrix $W \in \R^{m \times d}$ is a tunable parameters, and
$u$ with $\|u\| \leq 1$ is fixed throughout training.
In this scenario we obtain a merged predictor by simply summing hidden weight
matrices of neural networks trained on different tasks.
The intuition behind the argument in this case is that for an input $x$, the
length $\|W_t' x\|$ must be small because rows of $W_t'$ lie in the span on
$(x_i')_i$ meanwhile each of these points is sufficiently different from $x$.
We formalize this in the following lemma, shown in \Cref{proof:lem:hidden-null-space}, which applies to \ac{GD} iterates:
\begin{lemma}
  \label{lem:hidden-null-space}
  Suppose that inputs $(x_i)_i$ and $x'$ are such that
  $
    \max_i\abs{\ip{x_i, x'}} \leq \ve
  $.
  Suppose that $W_t$ is a weight matrix of a shallow neural network obtained by running \ac{GD} for $t$ steps given $(x_i, y_i)_{i=1}^n$.
  Assume that the loss function is $1$-Lipschitz (e.g. logistic loss).
  Then, for any $j \in [m], t \in \mathbb{N}$,
  \begin{align*}
    &\|W_t x'\| \leq
    \|W_0 x'\| \, (1-\eta \lambda)^t + \ve \, \frac{1-(1-\eta \lambda)^t}{\lambda}\\
    \text{which implies} \qquad
    &|f_{\th_t + \th_t'}(x) - f_{\th_t}(x)|
\leq \|W_0' x\| \, (1-\eta \lambda)^t + \ve \, \frac{1-(1-\eta \lambda)^t}{\lambda}~.
  \end{align*}
\end{lemma}
\paragraph{Deep linear networks}
Next, we consider the predictor
\begin{align*}
  f_{\th}(x) = \ip{w_K, W_{K-1} \cdots W_1 x}
\end{align*}
where each weight matrix is trained by \ac{GF} dynamics as described in \cref{eq:GF-W}.
To show the desired result we exploit a technical result of \cite{arora2018optimization} which translates \ac{GF} dynamics for individual matrices into \emph{implicit} \ac{GF} dynamics for the \emph{end-to-end} vector
$
  w(t)\tp = w_K(t)\tp W_{K-1}(t) \cdots W_1(t)~.
$
The proof, given in \Cref{proof:thm:deep-linear-multitask-cor}, exploits the fact that $w(t)$ indeed lies in the span of inputs.
\begin{theorem}
  \label{thm:deep-linear-multitask-cor}
  Assume that weight matrices are initialized such that
  for all $k \in [K-1]$
  \begin{align*}
    W_{k+1}(0)\tp W_{k+1}(0) = W_{k}(0) W_{k}(0)\tp~.
  \end{align*}
  Then, for any $t \geq 0$ and any input $x$ such that $\max_i \abs{\ip{x_i', x}} \leq \ve$ we have
  \begin{align*}
  |f_{\th(t)+\th'(t)}(x) - f_{\th(t)}(x)|
  \leq
  |f_{\th(0)}(x)| A_1 \, e^{- \lambda K  t}
  +
  A_2 \, \ve
\end{align*}
where terms $A_1, A_2, B$ depend only on initialization, $\lambda$, and $K$ (see
\Cref{thm:deep-linear-multitask} for precise constants).
\end{theorem}
\Cref{thm:deep-linear-multitask-cor} gives the bound on the gap of the same form as in the shallow and linear case, that is an exponentially decaying term that arises because of the weight decay, and an $\ve$-dependent term which captures similarity of inputs in different tasks.
Unlike the previous cases, the theorem assumes a particular initialization, which is inherited from \citet{arora2018optimization}, which is  benign and it is satisfied with high probability when entries of weight matrices are sampled from some symmetric distribution.

\section{Experiments}
\label{sec:experiments}
\paragraph{Norm and rank structure}
\label{sec:norm-rank-exps}
In this section we aim to investigate whether some of the biases discussed in
\Cref{sec:stationary} extend do very deep neural networks.  In particular, we
aim to verify whether norm preservation and low stable-ranks can be found in
some (smaller) LLMs.  Even though assumptions of \Cref{sec:stationary} might not
be satisfied, we still find that in several LLMs we observe low stable-ranks (drastically smaller than the dimension of the matrix), while norm preservation appears in most of the layers (MLP and attention ones). In these experiments we look at publicly available pre-trained BERT with
$\approx $ 110M parameters, GPT2-Large with $\approx$ 774M
, GPT2-XL with $\approx$ 1.5B , RoBERTa with $\approx$
125M , Phi-2 with $\approx$ 2.8B , GPT-J with $\approx$ 6B (we used Hugging Face ``Transformers'' library
  \citep{wolf-etal-2020-transformers}).
All experiments are performed on a computing platform with Nvidia A$100$ GPU.
  
  Each of these `transformer' models consists of a sequence of a so-called \emph{encoder} layers, where each encoder layer consists of a
  \emph{self-attention layer} followed by a fully-connected neural network. Self-attention layer is given by a matrix-to-matrix function $Q \mapsto \text{softmax}\big( Q K^T / \sqrt{\text{columns}(K)} \big) V$ 
 where $(Q,V)$ are parameter matrices and softmax is taken row-wise.
 Each self-attention layer is followed by a feed-forward fully connected neural network consisting of two linear layers with a non-linear activation function (e.g., ReLU or a smooth activation function) in between. In the context of transformer architecture, $Q,K,V$ are known as query, key, and value matrices.
 In \Cref{sec:experiments-appendix} we provide a table
  \Cref{tab:transformers} that summarizes architectural details of these models.
  
  While some of the results are given in \Cref{sec:experiments-appendix}, here in \Cref{fig:phi-2} we provide results for two largest neural networks.   In these plots, Q, K, V, O, and QKV denote Query, Key, Value, Output, and the concatenation of Query, Key, Value matrices of the attention layer, respectively. Weight matrices of MLP layers are denoted by M1 and M2.

The plots show that the Frobenius norms of QKV, M1, and M2 matrices are generally in the same order. Although this might appear puzzling at first, it is in fact consistent with \Cref{lem:norm-preservation}: the lemma is a statement about norms of layer weights when the output of the network can be written as a product of those layers. To apply the lemma to a transformer architecture, we should consider the QKV matrix as a layer weight instead of the individual attention matrices. All weight matrices generally have small stable-ranks, although the values can be different for different parameter type and layer. 

There are two observations for which we currently have no explanation: (1) The Frobenius norms of the same parameter type (even Q, K, V matrices) remain largely unchanged across layers. (2) Even though \Cref{lem:norm-preservation} suggests the O matrix should have similar Frobenius norms as QKV, M1, and M2 matrices, the plots show quite different values for norms of O matrices.
\newcommand{\figsize}{0.4}
\begin{figure}[H]
  \center
    \includegraphics[width=\figsize\linewidth]{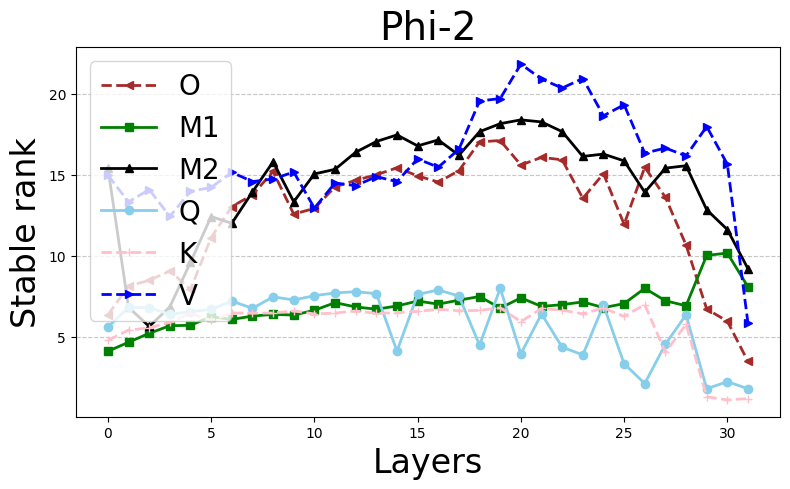}
    \includegraphics[width=\figsize\linewidth]{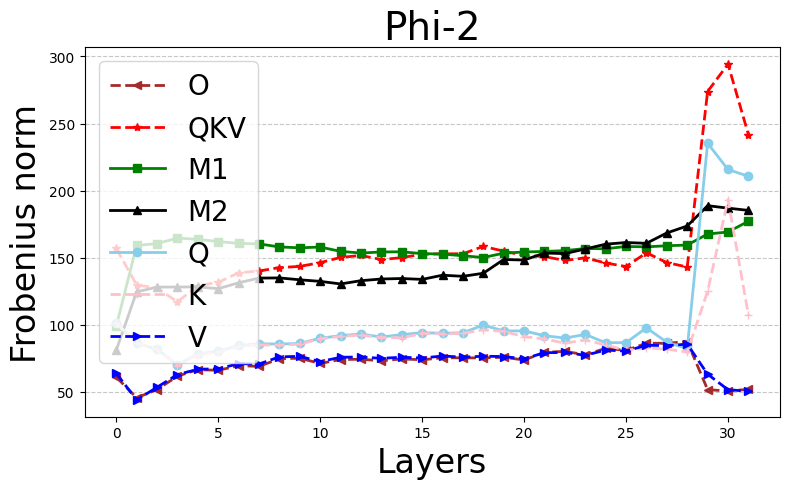}
    \includegraphics[width=\figsize\linewidth]{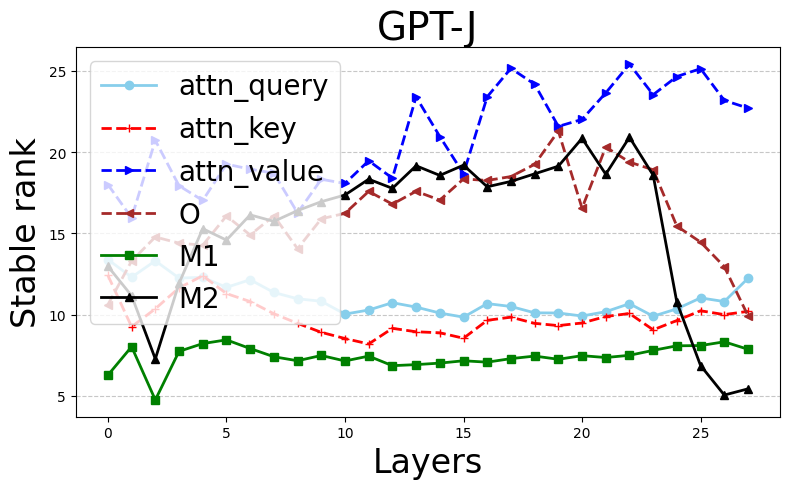}
    \includegraphics[width=\figsize\linewidth]{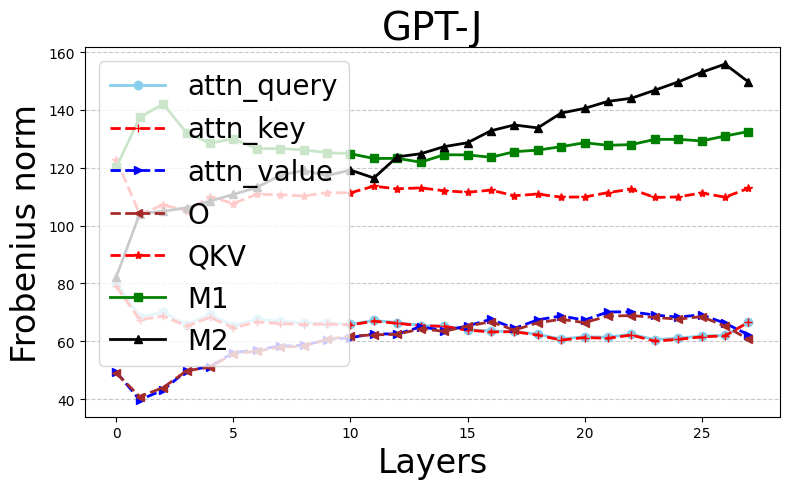}
\caption{Stable ranks and Frobenius norms of different weight matrices in pretrained Phi-2 (first row) and GPT-J (second row) models.} 
\label{fig:phi-2}
\end{figure}
Finally, \Cref{fig:rank-l2} shows that even if number of layers is small and network has a large number of classification mistakes, the average inverse stable-rank grows as weight decay parameter $\lambda$ increases. In this experiment, 1000 inputs are drawn independently from $\cN(0,I_d)$ with $d=100$ and then projected onto a unit sphere.
\begin{figure}[H]
  \center
    \includegraphics[width=\figsize\linewidth]{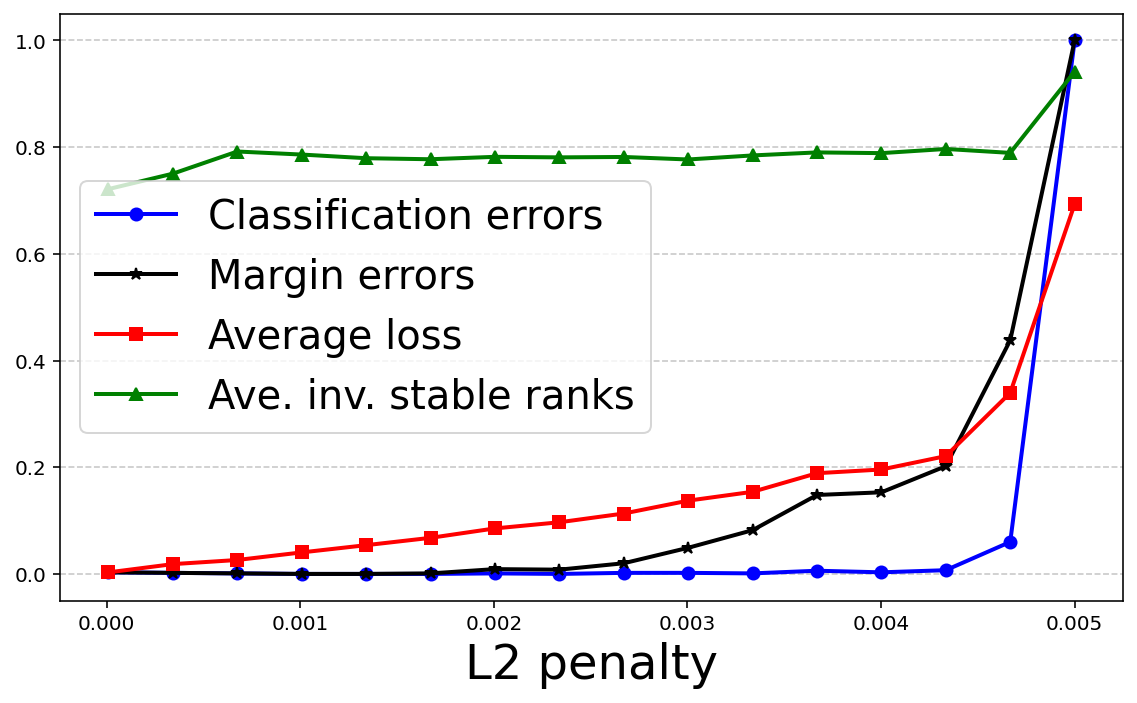}
    \includegraphics[width=\figsize\linewidth]{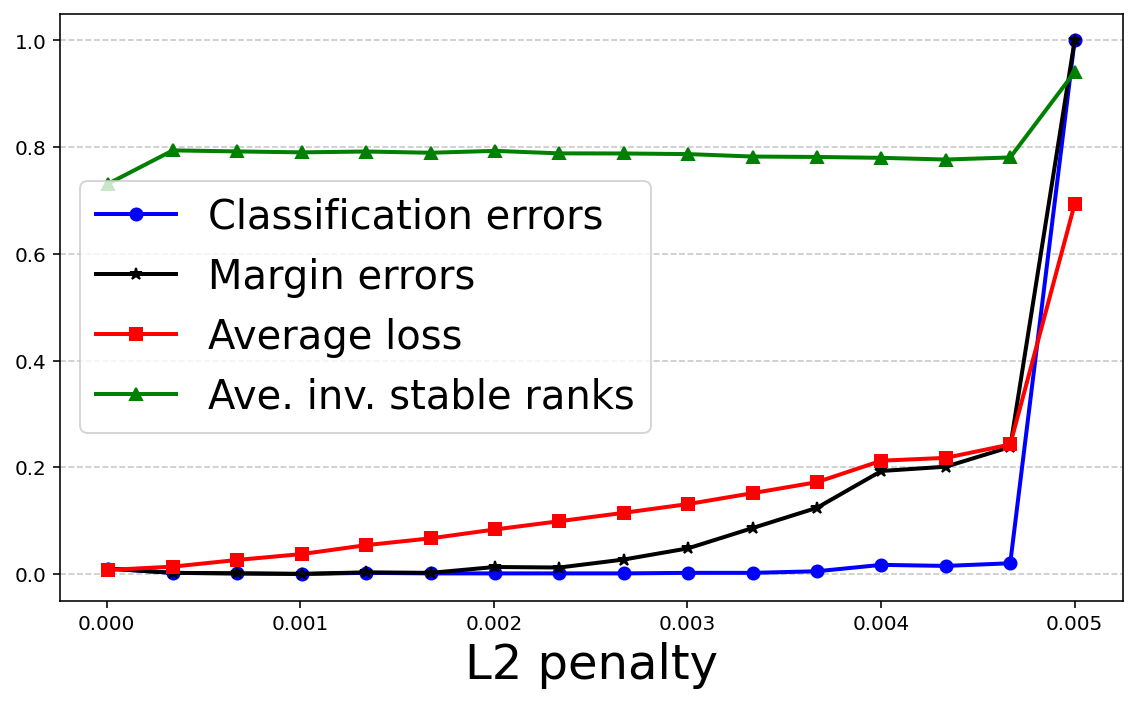}
\caption{Low rank induced by weight decay.} 
\label{fig:rank-l2}
\end{figure}
Given an input $x_i$, the label $y_i \in \{-1,+1\}$ is generated as
$y_i = 2 \, \ind\pr{1/(1+e^{-f^{\star}(x)}) \geq 1/2} - 1$ where the labeling
function is $f^{\star}(x) = \sin(10 W\tp x)$ (left plot) or $f^{\star}(x) = \sin(100 W\tp x)$ (right plot), and vector $W$ is
drawn from $\cN(0,I_d)$ and fixed throughout the experiment. For each value of $\lambda$, we use SGD with weight decay to train a network with two hidden layers (so, $K=3$), each of width 10. The number of epochs is 5000. \Cref{fig:rank-l2} shows the average inverse stable-rank of the final solution, along with its classification error, margin error (number of points with $y_i f_\theta(x_i) < 1$), and average loss. The plot shows that the average inverse stable-rank increases as $\lambda$ increases, even though the network might have large errors.
\paragraph{Model merging}
\label{sec:experiments:merging}

In our experiments we aim to verify four hypotheses:
(1) Training two neural networks on \emph{different tasks} (with nearly orthogonal
  inputs), and summing the weights results in a combined neural network that
  performs nearly as well on each of the tasks. Here the performance is
  understood in terms of the training loss;
(2) In a contrast, training two neural networks on the \emph{same task} and
  performing the merging as discussed above leads to the combined predictor that
  performs poorly on both tasks;
(3) This is behavior is enabled by the weight decay;
(4) Using weight decay leads to a low stable rank of weight matrices.

\emph{Data.}
The first set of experiments is performed on synthetic data, constructed as follows:
In case of independent tasks, inputs for the first task are drawn from isotropic
Gaussian $\cN(0,\Sigma_1)$, while for the second task inputs are drawn from
$\cN(0,\Sigma_2)$ where $\ip{\Sigma_1, \Sigma_2} = 0$ (this corresponds to
scenario $\ve = 0$ in \Cref{sec:merging}).
Given an input $x_i$, the label $y_i \in \{-1,+1\}$ is generated as
$y_i = 2 \, \ind\pr{1/(1+e^{-f^{\star}(x)}) \geq 1/2} - 1$ where the labeling
function $f^{\star}(x) = \sin(W\tp x)$.  For each task, vector $W$ is
drawn from $\cN(0,I_d)$ and fixed throughout the experiment.
In all experiments in this section inputs are normalized to lie on a unit sphere.
We achieve similar observations also in case of a real data, see \Cref{sec:mlp-more-results}.

\emph{Model and training.}
In all the experiments Fully-connected \ac{ReLU} neural network with inputs
$d=100$, two hidden layers of sizes $(1000, 100)$, and a scalar output.In each experiment on synthetic data, models are trained by \ac{GD} with step size $\eta = 1$ over
$10^5$ steps.  In all the experiments excepts the one where weight decay varies,
weight decay parameter $\lambda$ is set to $10^{-4}$.
On Fashion MNIST dataset, step size is set as $\eta=0.1$ while $\lambda = 10^{-3}$.
All experiments are repeated on $10$ random draws of the sample, and we report standard deviations in all plots.
\begin{figure}
  \center
  \includegraphics[width=0.36\linewidth]{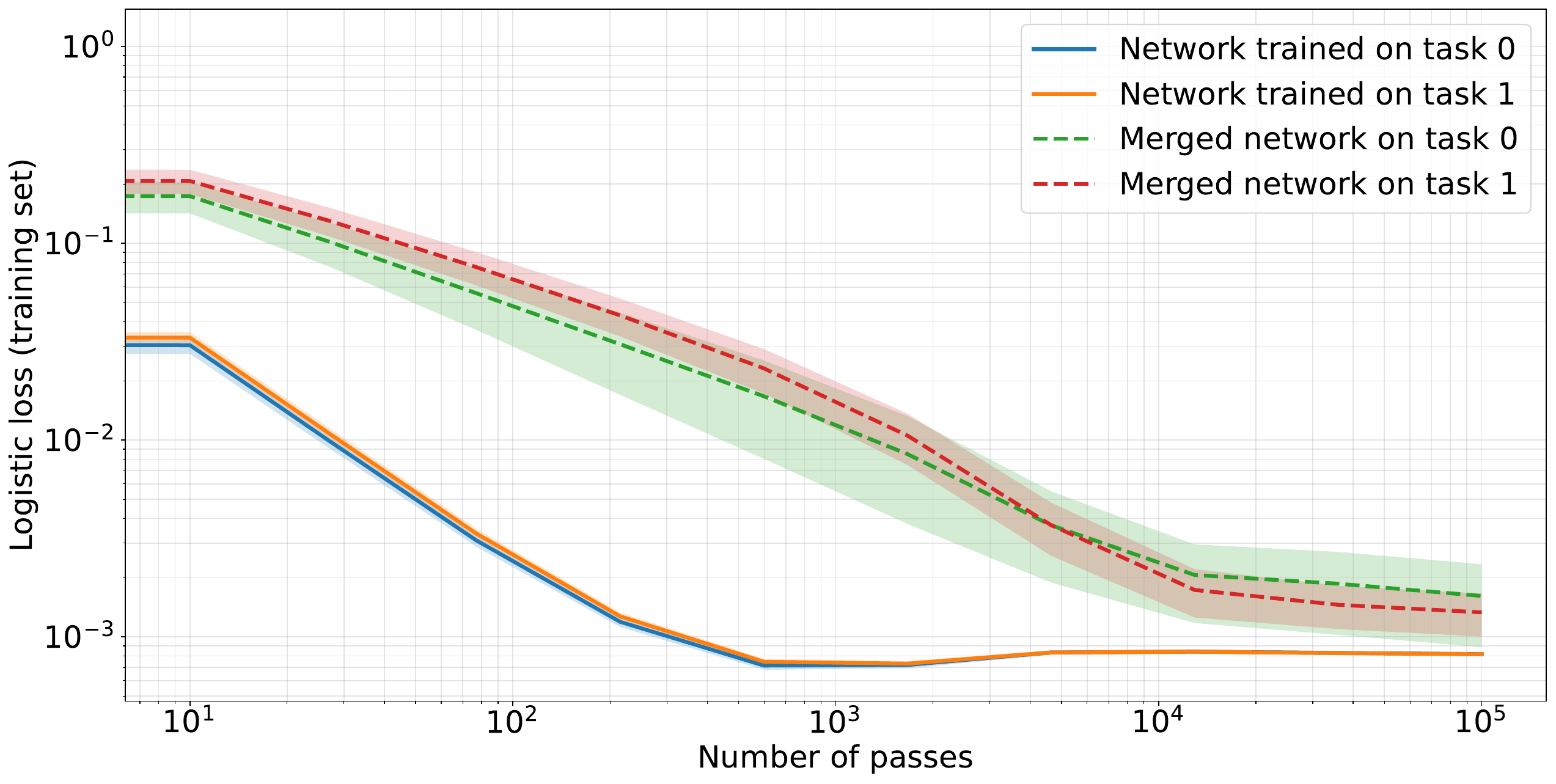}
  \includegraphics[width=0.36\linewidth]{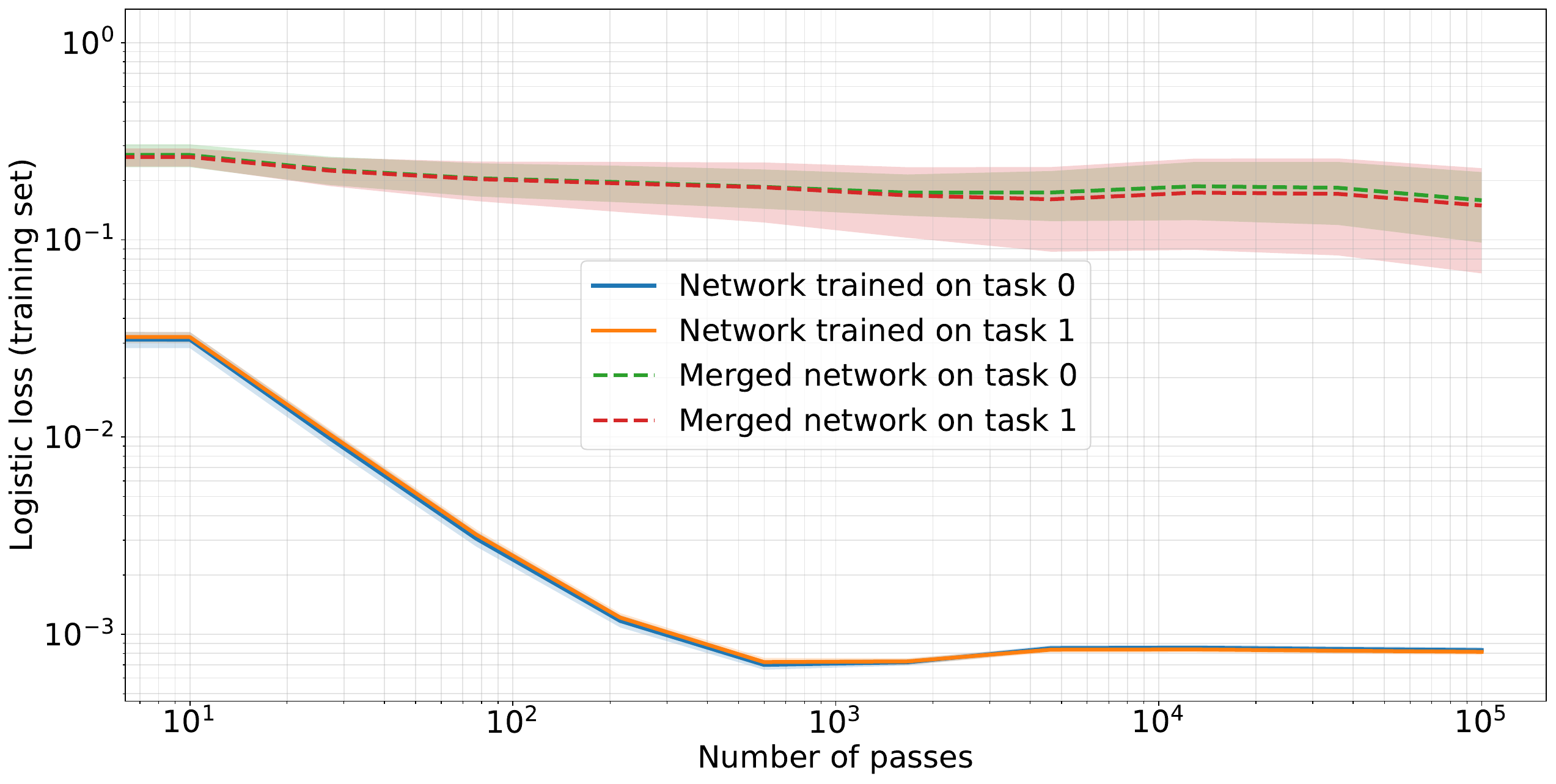}
  \includegraphics[width=0.36\linewidth]{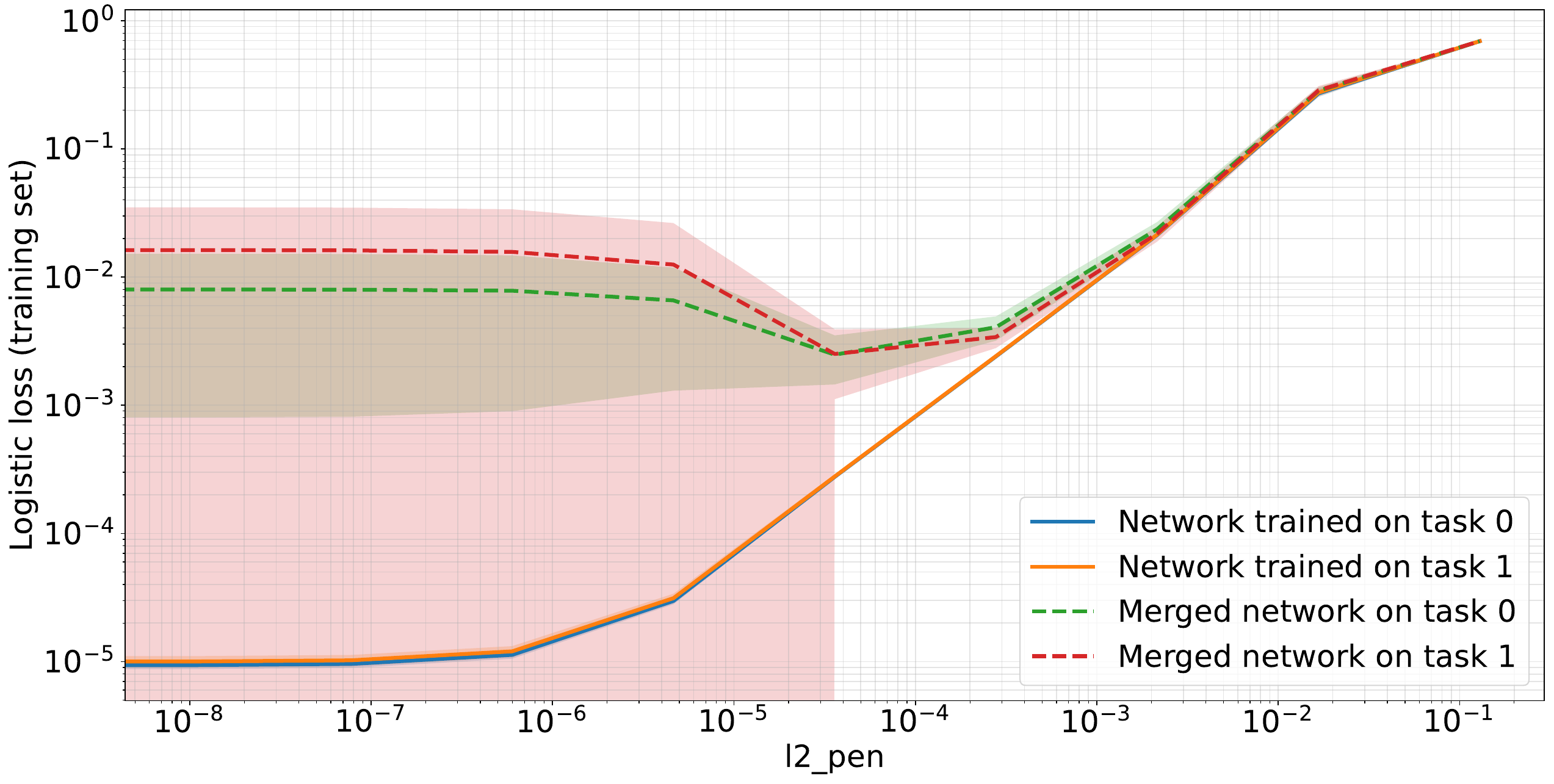}
  \includegraphics[width=0.36\linewidth]{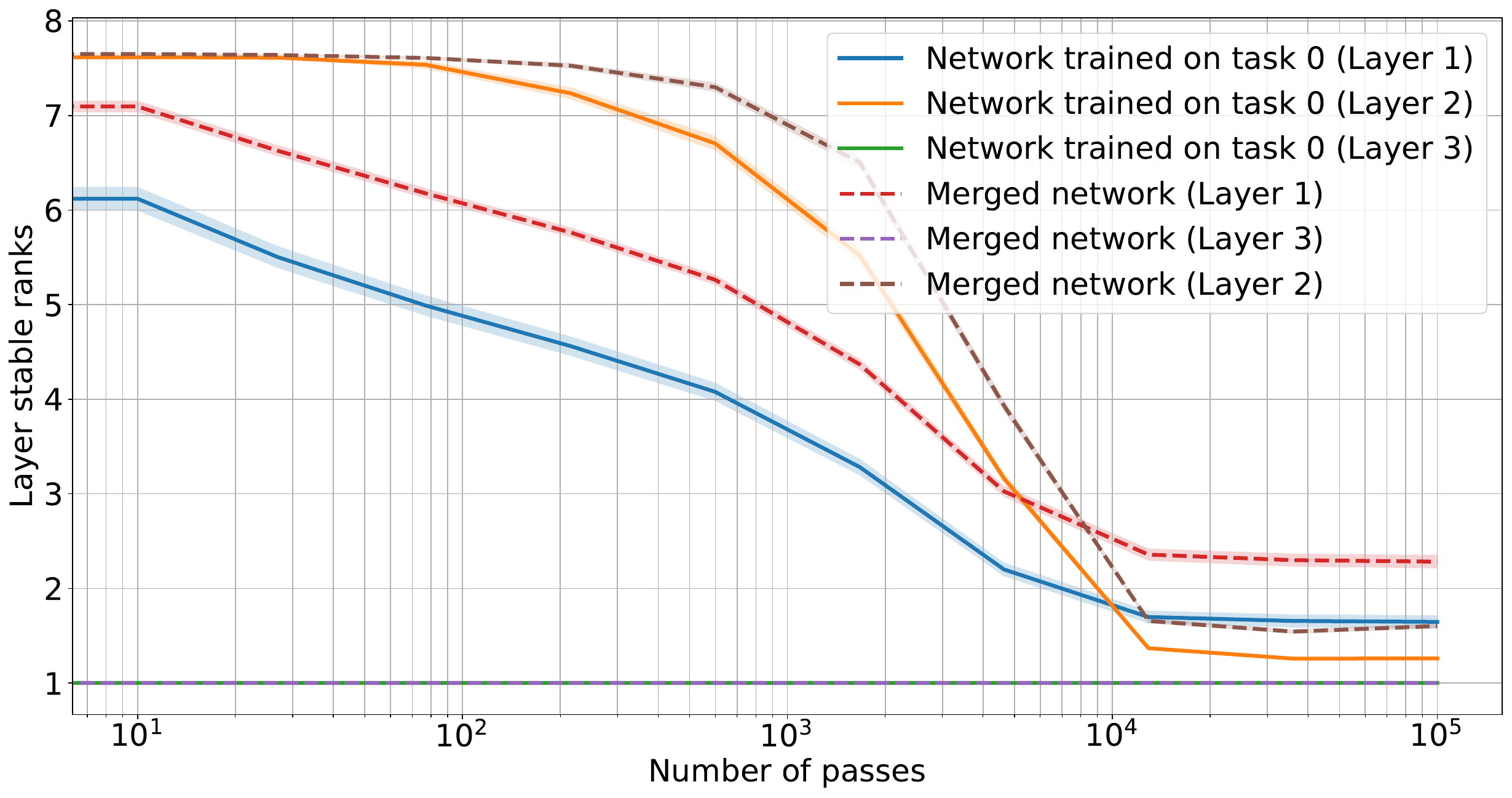}
  \caption{First row: Training neural networks on different tasks (orthogonal inputs) (left) vs. the same task (right) and
    merging the parameters by adding weight matrices. The resulting network
    performs well on different tasks after sufficiently many iterations, while given the same task, it does not.
    Second row: this effect manifests
    when weight decay strength is sufficiently large (left).
    Stable rank of each weight matrix converges to a small value, while stable rank of the merged 
    network matches stable ranks of individual networks (right).
    }
  \label{fig:t-vs-loss-different-tasks-synth}
\end{figure}

\emph{Discussion.}
On both datasets, we consistently observe that summing
weight matrices originating from different tasks, enables small logistic loss for merged models trained on all tasks after sufficiently long training (first row in \Cref{fig:t-vs-loss-different-tasks-synth}).
This seems to be in part enabled by orthogonality of inputs since, in contrast,
when inputs are not orthogonal and labels (or labeling functions) are different,
merged models do not get close in performance to distinct task-specific models.
Another
component that enables this gap is weight decay: In the second row of
\Cref{fig:t-vs-loss-different-tasks-synth} we observe that when
weight decay strength is not sufficient, the gap between losses of
merged and original models is substantial.  Finally, we observe that the stable
rank converges to a value much smaller than the actual rank at the stage when
performance of merged model actually gets close to the performance of original
model.  This is another observation in support of our hypothesis that low-rank bias is
crucial for model merging.

In this paper we focus on the training error, but one might wonder whether similar message about model merging carries over to \emph{generalization}, or preserving performance on the unseen but similarly distributed data. In \Cref{sec:mlp-more-results} we include experimental results confirming that model merging not only retains training errors of the original models, but also preserves their generalization ability.

\section{Conclusions}
In this work we examined the role of $L2$ regularization in training of deep
neural networks with logistic loss.  We investigated a surprising phenomenon:
merging two neural networks trained on sufficiently different tasks by simply
adding their respective weight matrices results in a predictor that performs
well on both tasks simultaneously.  As we attributed the explanation of this to
the low-rank bias arising in weight matrices, we also established that $L2$
regularization leads to weight matrices of a low stable rank.

These observations open up some interesting possibilities, especially in
multitask learning with large models, such as large language models, and
distributed optimization.

\bibliographystyle{plainnat}
\bibliography{learning}
 \newpage
\appendix
\clearpage

\section{Additional related work}
\label{sec:related}

A number of prior works have studied low-rank biases of neural networks in settings that are more restrictive than ours. 

\citet{jacot2022saddletosaddle} study low-rank biases of deep linear networks trained with GF and with near-zero variance initialization. In contrast, we consider deep linear networks optimized by GD, and we have no requirement on near-zero variance initialization. \citet{jacot2023implicitbiaslargedepth} study the low-rank phenomena in infinite-depth ReLU networks, while we study a finite-depth setting. \citet{GSGP-2024} show low-rank updates due to a small batch size. But these low-rank updates can eventually lead to a high-rank final solution if they are in different directions. Finally, \citet{YKWSVLMW-2024} empirically observe that a low-rank solution develops with weight-decay, and stable rank generally decreases during training. 

A number of works have considered shallow networks. \citet{phuonginductive} show convergence of a hidden weight matrix in a shallow neural network to a rank-one matrix when it is trained by \ac{GF} on \emph{orthogonally-separable data} (in the context of classification all inputs with a matching label satisfy $\ip{x_i, x_j} > 0$, and otherwise for a non-matching one). \citet{minearly} show convergence of stable rank to $2$ for shallow ReLU networks trained by \ac{GF} dynamics. Under mild assumptions on the inputs \citet{freiimplicit} established that the hidden weight matrix of a shallow  neural network convergences to the low-rank matrix (at most $2$). \citet{BPF-2022} study implicit bias of shallow networks (under orthogonal inputs) trained with GF and with small initialization. \citet{CYY-2024} show that for a two-layer ReLU network, training with SGD, weight decay, and minibatch leads to an approximately rank-two weight matrix. Our proof technique is different and is applicable to more general networks and the full-batch setting as well. 

\paragraph{Weight averaging}
Several prior works have considered building a final model by averaging weights of different models~\citep{Utans-1996,Izmailov-2018,Rame-2022,Wortsman-2022,Stojanovski-2022,Ilharco-2022,Ilharco-2023,chung2023parameter,Douillard-2023,Rame-2023,Rame2024WARMOT, rame2024warp}. To the best of our knowledge, prior works mostly consider settings where several models are trained on data from the same or similar tasks. Their main idea is that the randomization in data or parameter initialization leads to perturbations in the learned weights, and taking their average leads to more stable solutions. This is also known as ``linear mode connectivity"~\citep{frankle20a,Neyshabur2020WhatIB}. In contrast, we consider models trained on entirely different tasks, and we take the weights' summation instead of their average. Our explanation for the success of model merging is also very different and relies on the low-rank structures of the learned weights.  

\paragraph{Neural Collapse}
Low-rank bias is intimately related to a so-called \emph{Neural Collapse (NC)},
which is a form of clustering within the feature space of a neural network~\citep{papyan2020prevalence}.
NC hypothesis suggests that the network refines its representations so that inputs belonging to the same class (in context of classification) are pulled closer together, forming distinct clusters. Simultaneously, the network pushes the cluster centers (class means) away from each other, maximizing their separation.

Some of the
existing results are mostly in the Unconstrained Feature Model (UFM) where only the last-layer features and the output of linear classifier are learnable. This is in fact equivalent to learning a shallow linear network. Most studies show that the global optima of the loss function satisfies the conditions of neural collapse, without studying the dynamics of gradient descent.

\citet{yang2022inducing} demonstrated that in the Universal Feature Manifold (UFM) setting, fixing the linear output layer to satisfy a specific condition induces neural collapse in the feature vectors that minimize the loss, even with imbalanced data.  This fixed output layer approach allows them to extend typical neural collapse results, which often assume balanced data, to the more challenging imbalanced case.

\citet{RBF-2022} showed that for ReLU deep networks trained on balanced datasets using gradient flow with weight decay and a squared loss, critical points satisfying a "Symmetric Quasi-interpolation" assumption also exhibit neural collapse.  This assumption, which posits the existence of a classifier whose output depends only on the class label and not the specific data point, is presented as a key condition for their result.

Notably, model merging discussed here  differs from \emph{weight
averaging}~\citep{Utans-1996}: weight averaging relies on learned weights
converging to vicinity of each other, so taking their average leads to a more
stable solution. In our case, learned weights are far from each other, and the
average weights often do not perform well in practice. As our theory suggests,
we should sum the weights up, which in fact leads to good performance.

\clearpage
\section{Proofs from \Cref{sec:stationary}}
\label{proof:sec:stationary}

\begin{proof}[Proof of \Cref{lem:deep-linear}]
  Let $w^{(i)\top}=w_{K}\tp W_{K-1}\dots W_{i}$. Therefore, $f_{\theta}(x)= \ip{w^{(i)}, W_{i-1} \dots W_{1} x}$.
  Let $r_i(\th) := -y_i \ell'(y_i f_\th(x_i)) / (n \lambda)$.
  By alignment, we have 
\begin{align*}
W_{1} &= w^{(2)}\sumin r_{i}(\theta) x_i\tp \,,\\
W_{2} &= w^{(3)}\sumin r_{i}(\theta) (W_{1} x_i)\tp \,,\\
\vdots \\
W_{K-1} &= w^{(K)}\sumin r_{i}(\theta) (W_{K-2}\dots W_{1} x_i)\tp \,,\\
w_{K} &= \sumin r_{i}(\theta) (W_{K-1}\dots W_{1} x_i)\tp \;.
\end{align*}
It's easy to see that all weight matrices above are rank-1. 
\end{proof}

\begin{proof}[Proof of \Cref{lem:norm-preservation}]
\cref{eq:alignment} holds also for the parameters of layer $k$: $\lambda W_k = -\nabla_{W_k} L(\theta)$. Therefore,
\begin{align*}
\lambda \|W_k \|_F^2 &= \lambda\,\tr(W_k W_k\tp)\\ 
&= -\tr(\nabla_{W_k} L(\theta)W_k\tp) \\ 
&= -\frac{1}{n} \sumin \ell'(y_i f_{\theta}(x_i)) y_i \tr (\nabla_{W_k} f_{\theta}(x_i) W_k\tp)  \\
&= -\frac{H^{K-k}}{n} \sumin  \ell'(y_i f_{\theta}(x_i)) y_i f_{\theta}(x_i) \,,
\end{align*}
where the last step holds by the fact that for $f_\theta$ locally Lipschitz and positively homogeneous, $H^{K-k} f_{\theta}(x) = \tr (\nabla_{W_k} f_{\theta}(x_i) W_k\tp)$ (see e.g. Lemma~9.2 of \citet{mjt_dlt}).
Given that the equation holds independently of layer index $k$, weight matrices all have the same Frobenius norm. 
\end{proof}

\begin{proof}[Proof of \Cref{lem:low-rank}]
    By the weighted AM-GM inequality, and the fact that activations are $H$-homogeneous, we get that
\[
|f_\theta(x)| \le  \prod_k \| W_k\|_2^{H^{K-k}} \le \left(\frac{1}{Z}\sum_{k}^K H^{K-k} \| W_k\|_2 \right)^{Z}~.
\]
assuming that $\|x\|\leq 1$.

On the other hand, given that the loss is logistic, from \Cref{lem:norm-preservation} we have that,
\begin{align*}
\lambda\| W_{k'}\|_F^2 &= -\frac{H^{K-k'}}{n}\sum_{i} \ell'(y_i f_\theta(x_i)) y_i f_\theta(x_i) \\
&\le  \frac{H^{K-k'}}{n}\sum_{i} |\ell'(y_i f_\theta(x_i))| \cdot \max_j |f_\theta(x_j)| \\
&\le H^{K-k'} L(\theta) \cdot \max_j |f_\theta(x_j)|\\ 
&\le H^{K-k'} L(\theta)\left(\frac{1}{Z}\sum_{k}^K H^{K-k} \| W_k\|_2 \right)^{Z} \;.
\end{align*}
Introduce
$B^2 = (-1/n) \sum_i  \ell'(y_i f_{\theta}(x_i)) y_i f_{\theta}(x_i)$, and so by \Cref{lem:norm-preservation} $\|W_k\|_F = \sqrt{H^{K-k} B^2 / \lambda}$.
Therefore,
\begin{align*}
  \frac{1}{Z} \sum_k (H^{K-k})^{3/2} \frac{\|W_k\|_2}{\|W_k\|_F}
  &=
    \frac{1}{Z} \sum_k (H^{K-k})^{3/2}
    \frac{\sqrt{\lambda} \|W_k\|_2}{\sqrt{H^{K-k} B^2}}\\
  &=
    \frac{\sqrt{\lambda}}{B} \cdot \frac{1}{Z} \sum_k H^{K-k} \|W_k\|_2\\
  &\geq
    \frac{\sqrt{\lambda}}{B} \cdot \left(\frac{\lambda\| W_{k}\|_F^2}{H^{K-k} L(\theta)} \right)^{1/Z} \tag{For any $k$}\\
  &=
    \frac{\sqrt{\lambda}}{B} \cdot \left(\frac{H^{K-k} B^2}{H^{K-k} L(\theta)} \right)^{1/Z}\\
  &= \sqrt{\lambda} \cdot \frac{B^{2/Z-1}}{L(\theta)^{1/Z}}\\
  &\geq
    \sqrt{\lambda} \cdot \frac{L(\theta)^{(2/Z-1)/4}}{L(\theta)^{1/Z}}\\
  &=
    \sqrt{\lambda} L(\theta)^{-(\frac14 + \frac{1}{2 Z})}
\end{align*}
where the last inequality requires
$$
  B^2= -\frac1n \sum_{i} \ell'(y_i f_\theta(x_i)) y_i f_\theta(x_i)\\
  \leq \sqrt{L(\theta)}
  $$
  using the fact that $-x \ell'(x) \leq \sqrt{\ell(x)}$ for the logistic loss, and Jensen's inequality.
In particular, assuming that for $K\geq 2, H \geq 1$, we have $2/Z-1 \leq 0$, and so $B \mapsto B^{2/Z-1}$ is non-increasing
and we have $B^{2/Z-1} \geq (\sqrt{L(\theta)})^{(2/Z-1)/2}$.
\end{proof}

\clearpage

\section{Proofs from \Cref{sec:merging}}
\label{proof:sec:merging}
\subsection{Proof of \Cref{lem:hidden-null-space}}
\label{proof:lem:hidden-null-space}

  Observe that \ac{GD} update rule is
  \begin{align*}
    W_{s+1} = (1-\eta \lambda) W_s - \eta \, \frac1n \sumin y_i \, \ell'(y_i \, f_{\th_s}(x_i))  \, D_{s,i} \, u \, x_i\tp
  \end{align*}
  where $D_{s,i} = \diag(\ind(W_s x_i > 0))$.
  Then
  \begin{align*}
    W_{s+1} x' = (1-\eta \lambda) W_s x' - \eta \, \frac1n \sumin y_i \, \ell'(y_i \, f_{\th_s}(x_i)) D_{s,i} \, u \ip{x_i, x'}
  \end{align*}
  which together with Cauchy-Schwartz inequality implies
  \begin{align*}
    \|W_{s+1} x'\|
    &\leq (1-\eta \lambda) \|W_s x'\| + \eta \, \frac1n \sumin |y_i \, \ell'(y_i \, f_{\th_s}(x_i))| \|D_{s,i}\|_2 \, \|u\| \abs{\ip{x_i, x'}}\\
    &\leq (1-\eta \lambda) \|W_s x'\| + \ve \, \eta~.
  \end{align*}
  Observe that relation $x_{s+1} \leq a_s x_s + b_s$
unwinds from $t$ to $t_0$
as
$x_t \leq x_{t_0} \prod_{k=t_0}^{t-1} a_k + \sum_{s=t_0}^{t-1} b_s \prod_{k=s+1}^{t-1} a_k$.
So,
\begin{align*}
  \|W_t x'\|
  &\leq \|W_0 x'\| \, (1-\eta \lambda)^t + \ve \, \eta \, \sum_{s=0}^{t-1} \prod_{k=s+1}^{t-1} (1-\eta \lambda)\\
  &= \|W_0 x'\| \, (1-\eta \lambda)^t + \ve \, \eta \, \sum_{s=0}^{t-1} (1-\eta \lambda)^{t-s-1}~.
\end{align*}
\qed

\subsection{Proof of the shallow neural network case}
\label{proof:thm:shallow}
First observe that
\begin{align*}
    |f_{\th_t + \th_t'}(x) - f_{\th_t}(x)|
  &\leq \abs{ \ip{u, (W_t x + W_t' x)_+} - \ip{u, (W_t x)_+}}\\
  &\leq \|(W_t x + W_t' x)_+ - (W_t x)_+\| \tag{Cauchy-Schwartz inequality}\\
  &\leq \|W_t' x\|
\end{align*}
where the last inequality comes by a basic fact about \ac{ReLU}s (see
  \Cref{sec:def}).
  Now, since $\ell$ is $1$-Lipschitz,
  \begin{align*}
    \ell\pr{y  f_{\th_t + \th_t'}(x)} - \ell\pr{y  f_{\th_t}(x)}
    &=
      \ell\pr{y  \ip{u, (W_t x + W_t' x)_+}} - \ell\pr{y  \ip{u, (W_t x)_+}}\\
    &\leq |f_{\th_t + \th_t'}(x) - f_{\th_t}(x)|\\
    &\leq \|W_t' x\|~.
  \end{align*}
    At this point, we apply losses over $(x_i, y_i)_i$ and
  average to have
  \begin{align*}
    L(\th_t + \th_t') \leq L(\th_t) + \frac1n \sumin \|W_t' x_i\|~.
  \end{align*}
  Now we  use \Cref{lem:hidden-null-space} to control $\|W_t' x_i\|$.
  \qed

\subsection{Proof of \Cref{thm:deep-linear-multitask-cor}}
\label{proof:thm:deep-linear-multitask-cor}
\Cref{thm:deep-linear-multitask-cor} is a direct corollary of the following theorem which we show in \Cref{proof:thm:deep-linear-multitask}:
\begin{theorem}
  \label{thm:deep-linear-multitask}
  Assume that weight matrices are initialized such that
  for all $k \in [K-1]$
  \begin{align*}
    W_{k+1}(0)\tp W_{k+1}(0) = W_{k}(0) W_{k}(0)\tp~.
  \end{align*}
  Let $\ell$ be a logistic loss and let $L(\th(0)) \leq C$.
  Then, for any $t \geq 0$ and any input $x'$ such that $\max_i \abs{\ip{x_i, x'}} \leq \ve$ we have
  \begin{align*}
  |\ip{w(t), x'}|
  \leq
  |\ip{w(0), x'}| A_1 \, e^{- \lambda K  t}
  +
  A_2 \, \ve
\end{align*}
where
\begin{align*}
  A_1 &= \exp\pr{\frac{B^{2 - \frac{2}{K}} (K-1) (1 +  \lambda K) C}{2 \lambda K}}~,\\
  A_2 &= 2  B^{2 - \frac{2}{K}} C \, A_1 \pr{\frac{1 - e^{- \lambda K t / 2}}{\lambda K}}~,\\
  B &= \|w(0)\|^2 + \frac{C}{\lambda}~.
\end{align*}
\end{theorem}
In the following for the end-to-end vector $w(t)\tp = w_K(t)\tp W_{K-1}(t) \cdots W_1(t)$ we use notation $L^1$ to denote its loss:
\begin{align*}
    L^1(w(t)) = \frac1n \sumin \ell(y_i  \ip{w(t), x_i})~.
\end{align*}
Note that $L^1(w(t)) =  L(\th(t))$.
Proof relies on the following crucial result connecting per-layer updates to
updates of the product matrix:
\begin{theorem}[{\citet[Theorem 1]{arora2018optimization}}]
  \label{thm:gf-linear-update}
  Assume that weight matrices are initialized (at time $t_0$) in such
  a way that they satisfy for all $k \in [K-1]$.
  \begin{align*}
    W_{k+1}(t_0)\tp W_{k+1}(t_0) = W_{k}(t_0) W_{k}(t_0)\tp
  \end{align*}
  Then, under dynamics with updates as in \cref{eq:GF-W} for the
  end-to-end matrix we have
  \begin{align*}
    \dot W(t) = -  \lambda K \cdot W(t)
    - \sum_{k=1}^K \pr{W(t) W(t)\tp}^{\frac{k-1}{K}}
    \cdot \nabla L^1(W) \cdot \pr{W(t)\tp W(t)}^{\frac{K-k}{K}}~.
  \end{align*}
\end{theorem}
\subsubsection{Proof of \Cref{thm:deep-linear-multitask}}
\label{proof:thm:deep-linear-multitask}
We start from adapting \Cref{thm:gf-linear-update} to our case to get $w(t)$ and then get an identity for $\ip{w(t), x'}$ by solving the resulting differential equation.
Once we get dependence on $\ve$ we must ensure that the remaining terms (arising from the solution to differential equation) are non-divergent, which we will do through the stationary point convergence argument.

\Cref{thm:gf-linear-update} gives us
\begin{align*}
  \dot w(t)\tp
  &= - \lambda K \cdot w(t)\tp
    -  \|w(t)\|^{\frac{2(K-1)}{K}} \cdot \nabla L^1(w(t))\tp\\
  &\qquad-  \sum_{k=1}^{K-1} \|w(t)\|^{\frac{2(k-1)}{K}} \cdot \nabla L^1(w(t))\tp \pr{w(t) w(t)\tp}^{\frac{K-k}{K}}\\
  &= - \lambda K \cdot w(t)\tp
    -  \|w(t)\|^{2 - \frac{2}{K}} \cdot \nabla L^1(w(t))\tp\\
  &\qquad-  \sum_{k=1}^{K-1} \|w(t)\|^{\frac{2(k-1)}{K} + \frac{2(K-k)}{K}} \cdot \nabla L^1(w(t))\tp \pr{\frac{w(t) w(t)\tp}{\|w(t)\|^2}}^{\frac{K-k}{K}}\\
  &= - \lambda K \cdot w(t)\tp
    -  \|w(t)\|^{2 - \frac{2}{K}} \cdot \nabla L^1(w(t))\tp\\
  &\qquad-  (K-1) \|w(t)\|^{2 - \frac{2}{K}} \cdot \ip{\nabla L^1(w(t)), w(t)} w(t)\tp~.
\end{align*}
Put another way,
\begin{align}
  \label{eq:linear-nets-upd}
  \dot w(t)
  = \pr{- \lambda K -  (K-1) \|w(t)\|^{2 - \frac{2}{K}} \, \ip{\nabla L^1(w(t)), w(t)}} \, w(t)
  -  \|w(t)\|^{2 - \frac{2}{K}} \, \nabla L^1(w(t))~.
\end{align}
Taking dot product with $x'$ gives
\begin{align}
  \label{eq:linear-nets-upd-x}
  \ip{\dot w(t), x'} = a(t) \, \ip{w(t), x'} + b(t)
\end{align}
where we introduce abbreviations
\begin{align*}
  a(t) &:= - \lambda K -  (K-1) \|w(t)\|^{2 - \frac{2}{K}} \ip{\nabla L^1(w(t)), w(t)}\\
  b(t) &:= - \|w(t)\|^{2 - \frac{2}{K}} \ip{\nabla L^1(w(t)), x'}~.
\end{align*}
Solving \cref{eq:linear-nets-upd-x} we get
\begin{align*}
  \ip{w(t), x'}
  &=
    \ip{w(t_0), x'} \, e^{\int_{t_0}^t a(s) \diff s}
    +
    \int_{t_0}^{t} b(s) \, e^{\int_s^t a(r) \diff r} \diff s~.
\end{align*}
Assuming for a moment that $\|w(t)\| \leq B$ (where $B$ will be determined
later) Cauchy-Schwartz inequality gives
\begin{align*}
  &a(t)
    \leq - \lambda K +  (K-1) B^{1 - \frac{1}{K}} \|w(t)\|^{1-\frac{1}{K}} |\ip{\nabla L^1(w(t)), w(t)}|~.
\end{align*}
On the other hand, using the fact that for logistic loss function $|\ell'(z)| \leq \ell(z)$,
\begin{align*}
  b(t)
  &=
     \|w(t)\|^{2 - \frac{2}{K}} \, \frac1n \sumin \ell'(y_i \ip{w(t), x_i}) y_i \ip{x_i, x'}\\
  &\leq
    \ve \cdot  B^{2 - \frac{2}{K}} \, \frac1n \sumin \ell(y_i \ip{w(t), x_i}) \nonumber\\
  &= \ve \cdot  B^{2 - \frac{2}{K}} \, L^1(w(t))~. \nonumber
\end{align*}
So, it is left to show that the term $\int_{t_0}^t a(s) \diff s$ does not diverge.
To this end, we show the following (with proof at the end of the section):
\begin{lemma}[Stationary point convergence for dynamics in \cref{eq:linear-nets-upd}]
  \label{lem:linear-nets-stationary}
  \begin{align*}
   (K-1) \int_s^t \|w(r)\|^{1-\frac{1}{K}} |\ip{\nabla L^1(w(r)), w(r)}| \diff r
  \leq
  \sqrt{ (K-1) (1 +  \lambda K) L^1(w(0)) \cdot (t-s)}~.
\end{align*}
\end{lemma}
Using this lemma to bound $\int a(s) \d s$ gives
\begin{align*}
  |\ip{w(t), x'}|
  &\leq
    |\ip{w(0), x'}| \, \underbrace{\exp\pr{- \lambda K \cdot t + B^{1 - \frac{1}{K}} \sqrt{ (K-1) (1 +  \lambda K) C \cdot t}}}_{(i)}\\
  &\qquad+
    \ve \cdot  B^{2 - \frac{2}{K}} C \, \underbrace{\int_{0}^{t} \exp\pr{- \lambda K \cdot (t-s) + B^{1 - \frac{1}{K}} \sqrt{ (K-1) (1 +  \lambda K) C \cdot (t-s)}} \diff s}_{(ii)}
\end{align*}
where we also assumed that $\sup_t L^1(w(t)) \leq C$.
At this point we will bound $(i)$ and $(ii)$ by using the fact that
\begin{align*}
  e^{-a t + b \sqrt{t}} \leq
  e^{\frac{b^2}{2 a}} \, e^{-a t/2} \qquad (a,b,t > 0)~.
\end{align*}
that comes from (choosing $p$ such that $\frac{b}{2 \sqrt{p}} = a/2$):
\begin{proposition}
  For any $a,b,t,p > 0$,
  \begin{align*}
    -a t + b \sqrt{t} \leq \pr{-a + \frac{b}{2 \sqrt{p}}} t  + \frac{b}{2} \sqrt{p}~.
  \end{align*}
\end{proposition}
In particular this gives
\begin{align*}
  (ii) \leq 2 \pr{\frac{1 - e^{- \lambda K t / 2}}{ \lambda K}} \exp\pr{\frac{B^{2 - \frac{2}{K}} (K-1) (1 +  \lambda K) C}{2 \lambda K}}
\end{align*}

As promised, the final bit is to give $B$.
Since objective is non-increasing
\begin{align*}
  \lambda \|w(t)\|^2 \leq \lambda \|w(t)\|^2 + L^1(w(t)) \leq \lambda \|w(0)\|^2 + L^1(w(0)) \qquad \implies \qquad
  B = \|w(0)\|^2 + \frac{L^1(w(0))}{\lambda}~.
\end{align*}

\begin{proof}[Proof of \Cref{lem:linear-nets-stationary}.]
Using the chain rule together with \cref{eq:linear-nets-upd} we have
\begin{align*}
  \frac{\d L^1(w(t))}{\d t}
  &= \ip{\nabla L^1(w(t)), \dot w(t)}\\
  &=
    - \lambda K \ip{\nabla L^1(w(t)), \dot w(t)} -  (K-1) \|w(t)\|^{2 - \frac{2}{K}} \, \ip{\nabla L^1(w(t)), w(t)}^2\\
    &-  \|w(t)\|^{2 - \frac{2}{K}} \, \|\nabla L^1(w(t))\|^2
\end{align*}
and so
\begin{align*}
  L^1(w(t)) - L^1(w(0))
  &=
  - \lambda K \int_0^t \ip{\nabla L^1(w(s)), \dot w(s)} \diff s\\
  &-  (K-1) \int_0^t \|w(s)\|^{2 - \frac{2}{K}} \, \ip{\nabla L^1(w(s)), w(s)}^2 \diff s\\
  &-  \int_0^t \|w(s)\|^{2 - \frac{2}{K}} \, \|\nabla L^1(w(s))\|^2 \diff s~.
\end{align*}
Note also that the chain rule gives
$
  \int_0^t \ip{\nabla L^1(w(s)), \dot w(s)} \diff s
  =
  L^1(w(t)) - L^1(w(0))
$
which gives
\begin{align*}
    & (K-1) \int_0^t \|w(s)\|^{2 - \frac{2}{K}} \, \ip{\nabla L^1(w(s)), w(s)}^2 \diff s
      +
       \int_0^t \|w(s)\|^{2 - \frac{2}{K}} \, \|\nabla L^1(w(s))\|^2 \diff s\\
    &\qquad= (1 +  \lambda K) \pr{L^1(w(0)) - L^1(w(t))}~.
  \end{align*}
  Applying Jensen's inequality completes the proof.
\end{proof}
\qed

\clearpage

\section{Additional empirical results from \Cref{sec:experiments}}
\label{sec:experiments-appendix}
\subsection{Transformer results}
\begin{figure}[H]
   \center
    \includegraphics[width=0.37\linewidth]{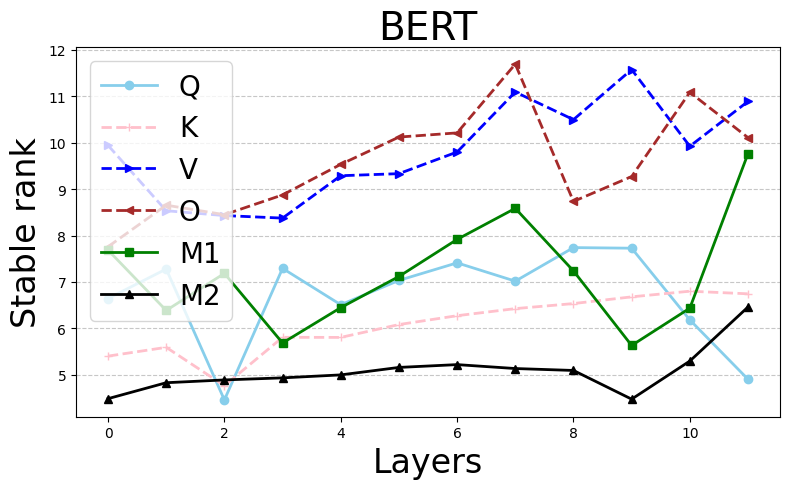}
    \includegraphics[width=0.37\linewidth]{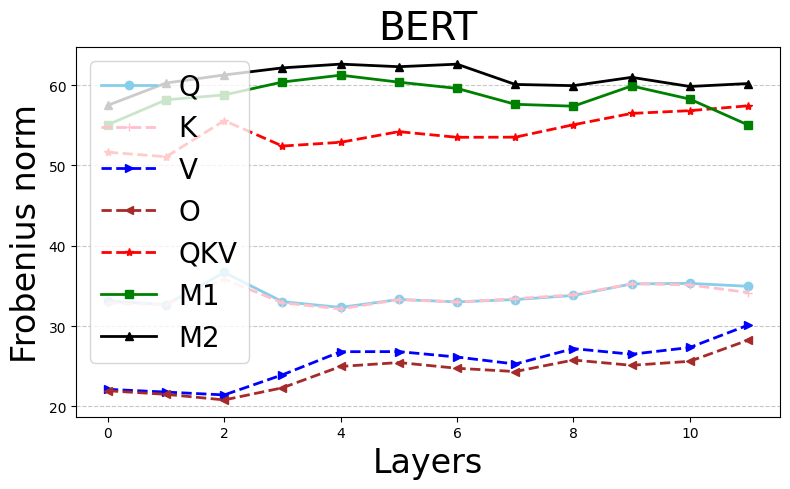}
    \includegraphics[width=0.37\linewidth]{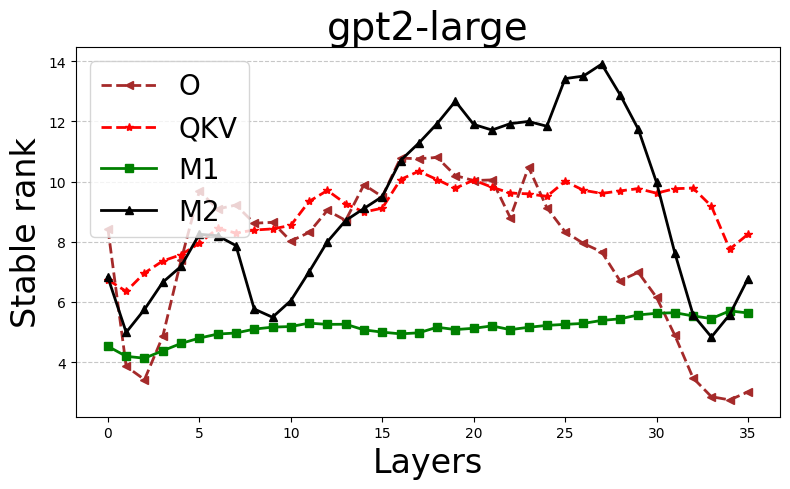}
    \includegraphics[width=0.37\linewidth]{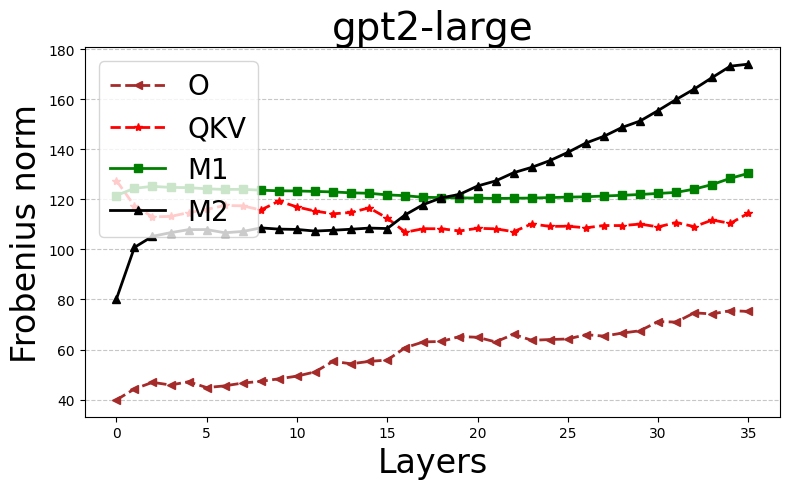}
    \includegraphics[width=0.37\linewidth]{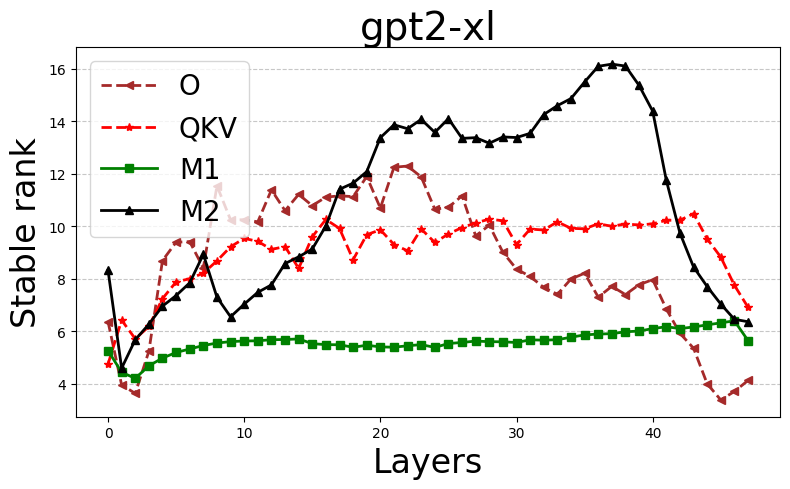}
    \includegraphics[width=0.37\linewidth]{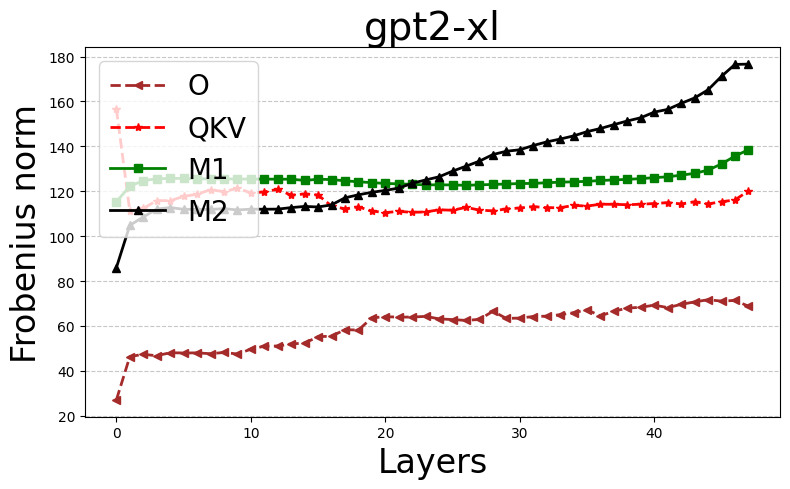}
    \includegraphics[width=0.37\linewidth]{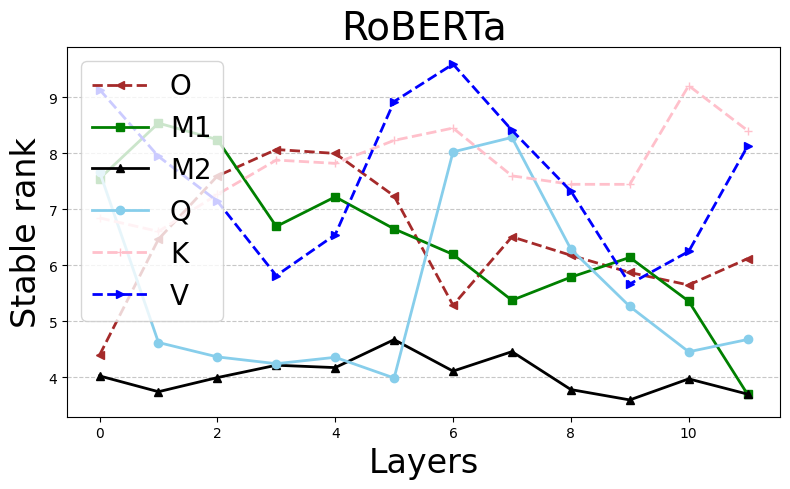}
    \includegraphics[width=0.37\linewidth]{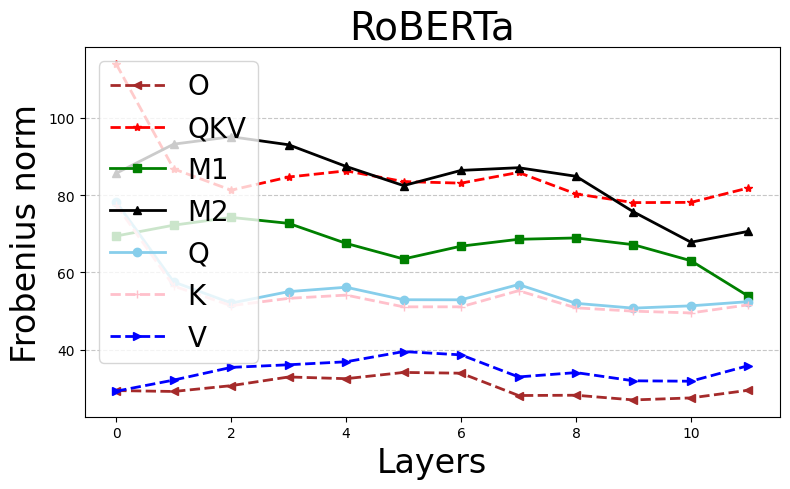}
\caption{Stable ranks and Frobenius norms of different weight matrices in pretrained BERT (first row), GPT2-Large (second row), RoBERTa (third row), and GPT2-XL (fourth row) model.} 
\label{fig:bert}
\end{figure}

\Cref{fig:bert} summarizes more measurements in addition to those in \Cref{sec:experiments}.
\begin{table}[h!]
\centering
\begin{tabular}{|c|c|c|c|}
\hline
\textbf{Model name} & \textbf{number of layers} & \textbf{model dimension} & \textbf{FF dimension} \\ \hline
BERT                & 12                & 768           & 3072           \\ \hline
RoBERTa & 12 & 768 & 3072            \\ \hline
 GPT2-Large & 36 & 1280 & 5120 \\ \hline
 GPT2-XL & 48 & 1600 & 6400 \\ \hline
 Phi-2 & 32 & 2560 & 10240 \\ \hline
 GPT-J & 28 & 4096 & 16384 \\ \hline
\end{tabular}
\caption{Specification of Transformer models.}
\label{tab:transformers}
\end{table}
 In \Cref{tab:transformers} model dimension refers to dimensionality of self-attention weight matrices, while FF dimension refers to the size of the hidden layer of the feedforward network.  

\subsection{MLP results}
\label{sec:mlp-more-results}
\begin{figure}[H]
\center
\includegraphics[width=0.49\linewidth]{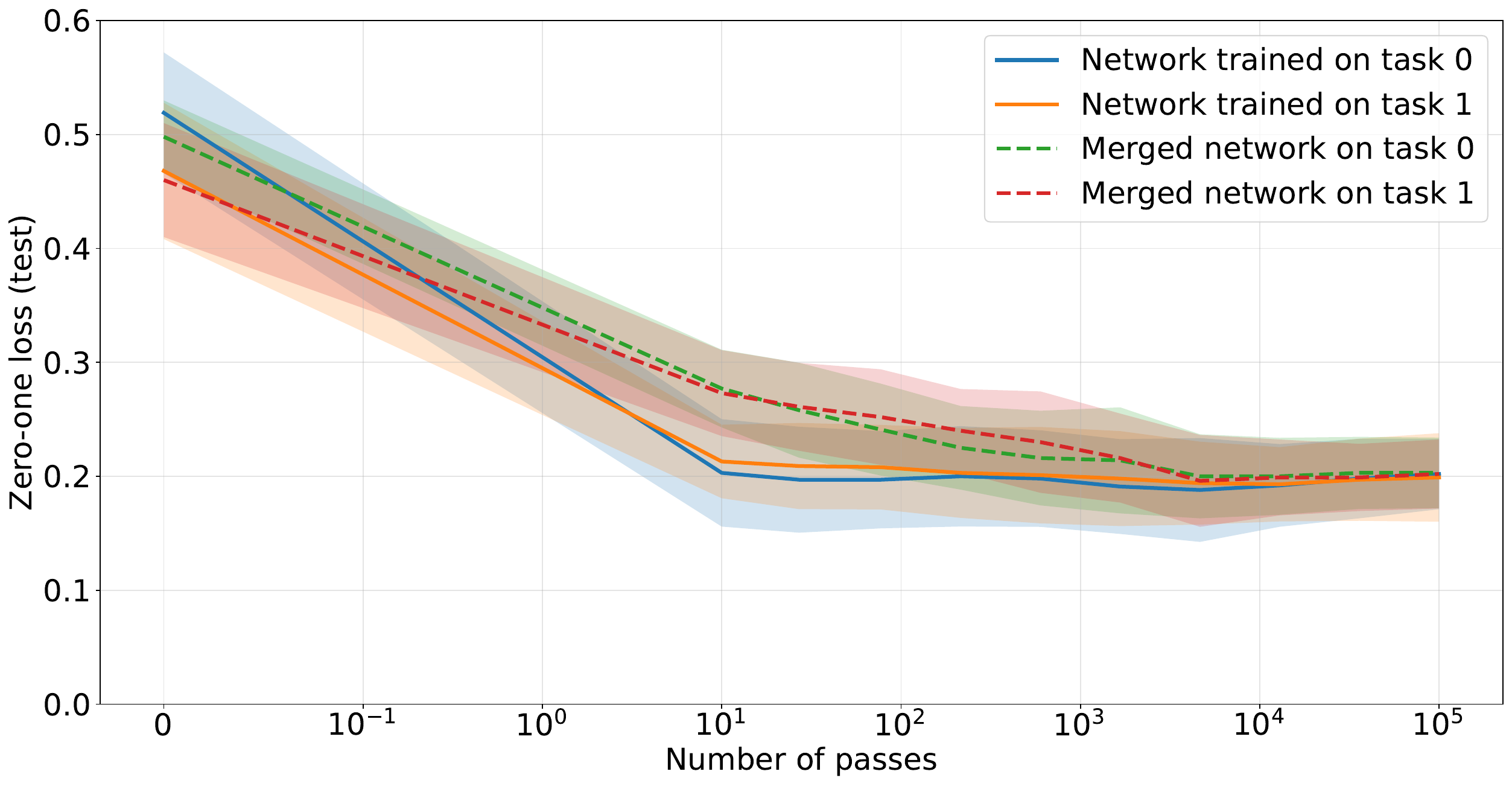}
\includegraphics[width=0.49\linewidth]{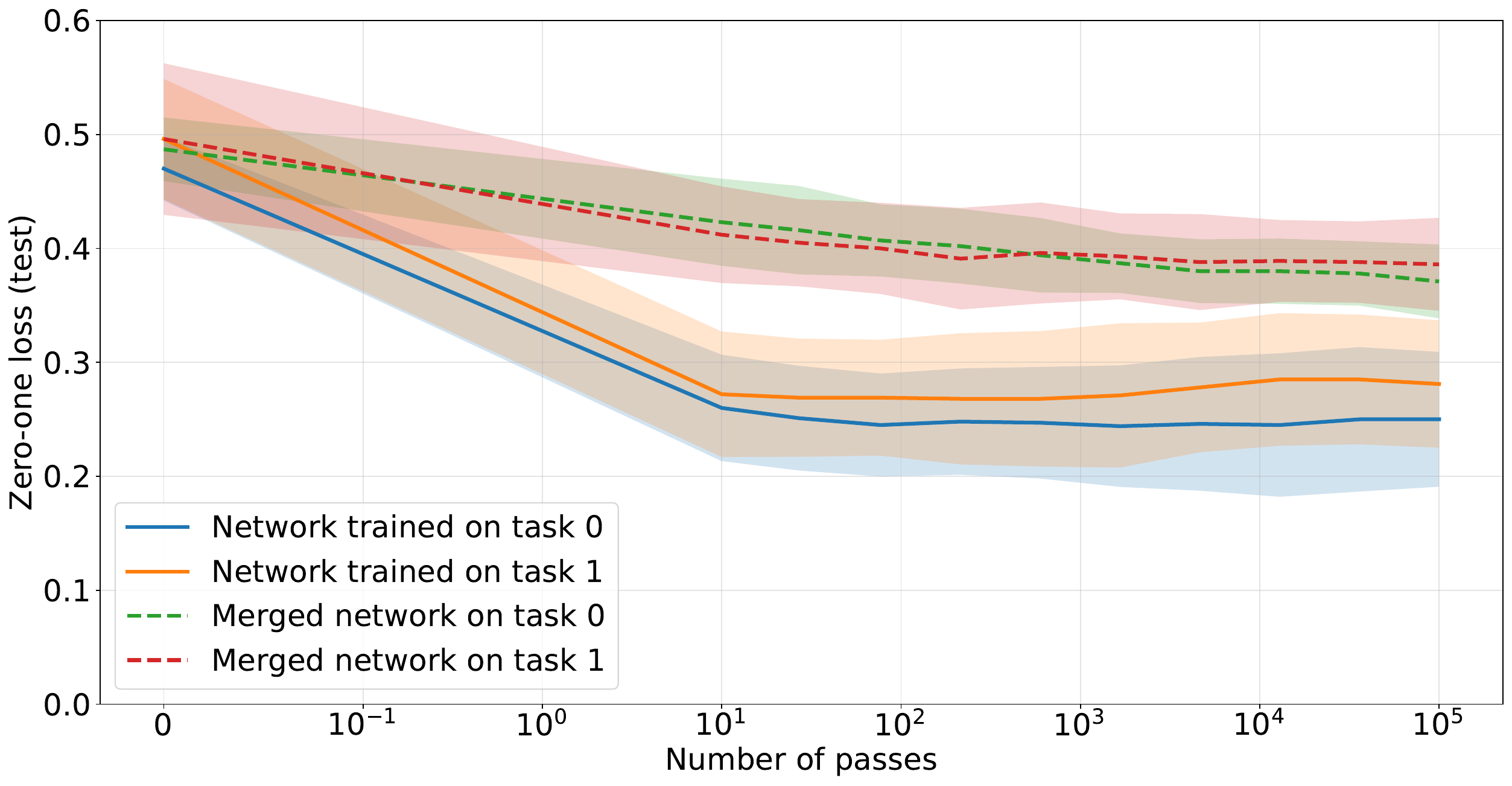}
\label{fig:synth-test-error}
\caption{Misclassification test error (on the held out sample) for merged models (left figure corresponds to orthogonal datasets, while right to the same source), in the same setting as in \Cref{fig:t-vs-loss-different-tasks-synth}.}
\end{figure}

We first include \Cref{fig:synth-test-error} which presents results for misclassification test error (on the held out sample) for merged models (left figure corresponds to orthogonal datasets, while right to the same source), in the same setting as in \Cref{fig:t-vs-loss-different-tasks-synth}.
Interestingly, merged models perform just as well as the original ones in term of the test error, whereas in the ``same task'' case the gap is apparent.
This indicates that model merging not only retains training errors of the original models, but also preserves generalization ability.

The second set of experiments is performed on `Fashion MNIST'
dataset~\citep{xiao2017fashion}, which contains grayscale images of 28x28 pixels
each, representing clothing items from 10 different categories.  We adapt this
dataset with sample size $10$ for binary classification by grouping first 5 classes into class 0, and
remaining into class 1.
When we consider two different tasks, we append 784-dimensional zero vector for the first task, and prepend in case of the second task.
This way, inputs from different tasks remain orthogonal, while the length of each input is preserved.
Finally, in case of task one binary labels are preserved as is, while for the second task labels are inverted.
\begin{figure}[H]
  \includegraphics[width=0.5\linewidth]{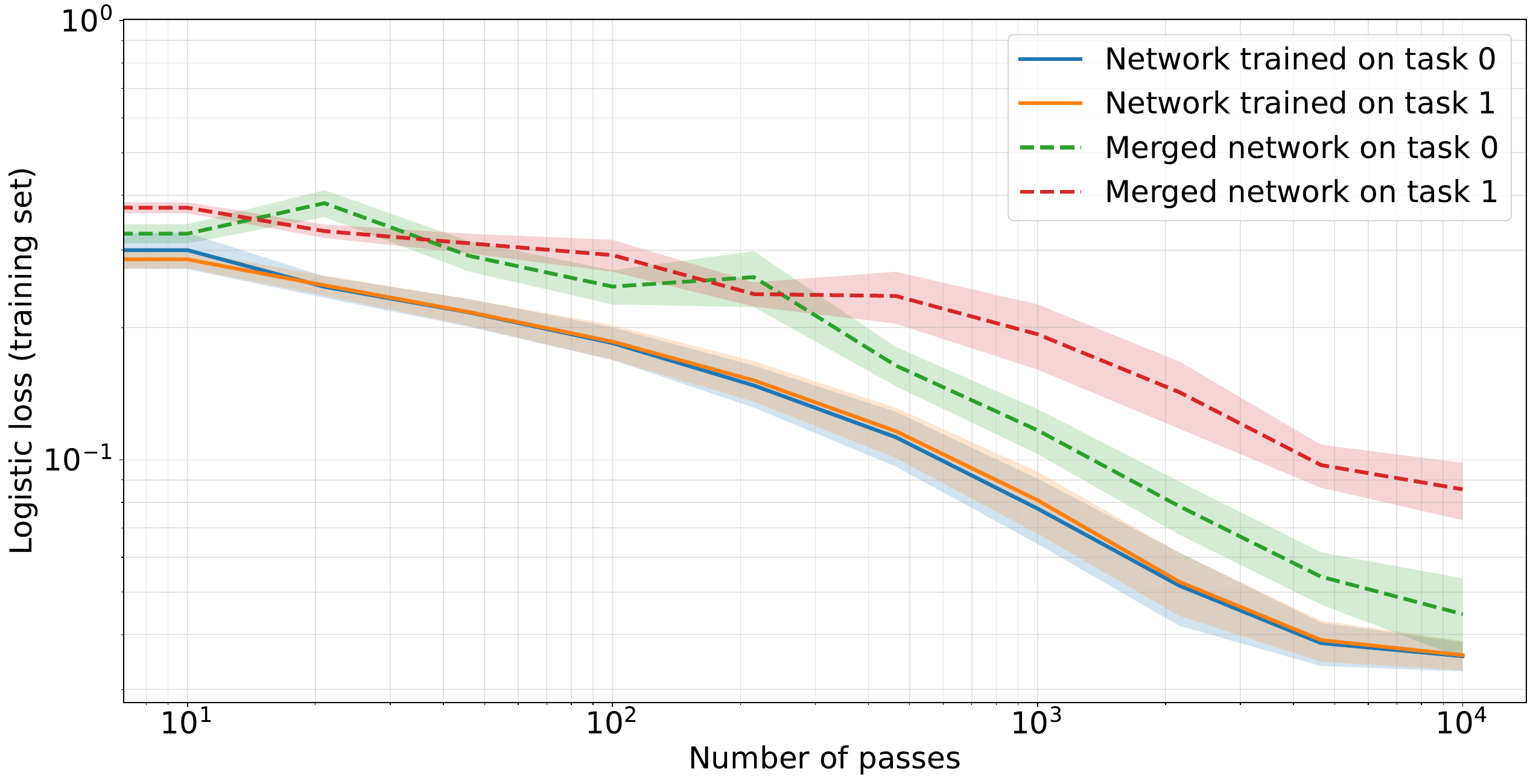}
  \includegraphics[width=0.5\linewidth]{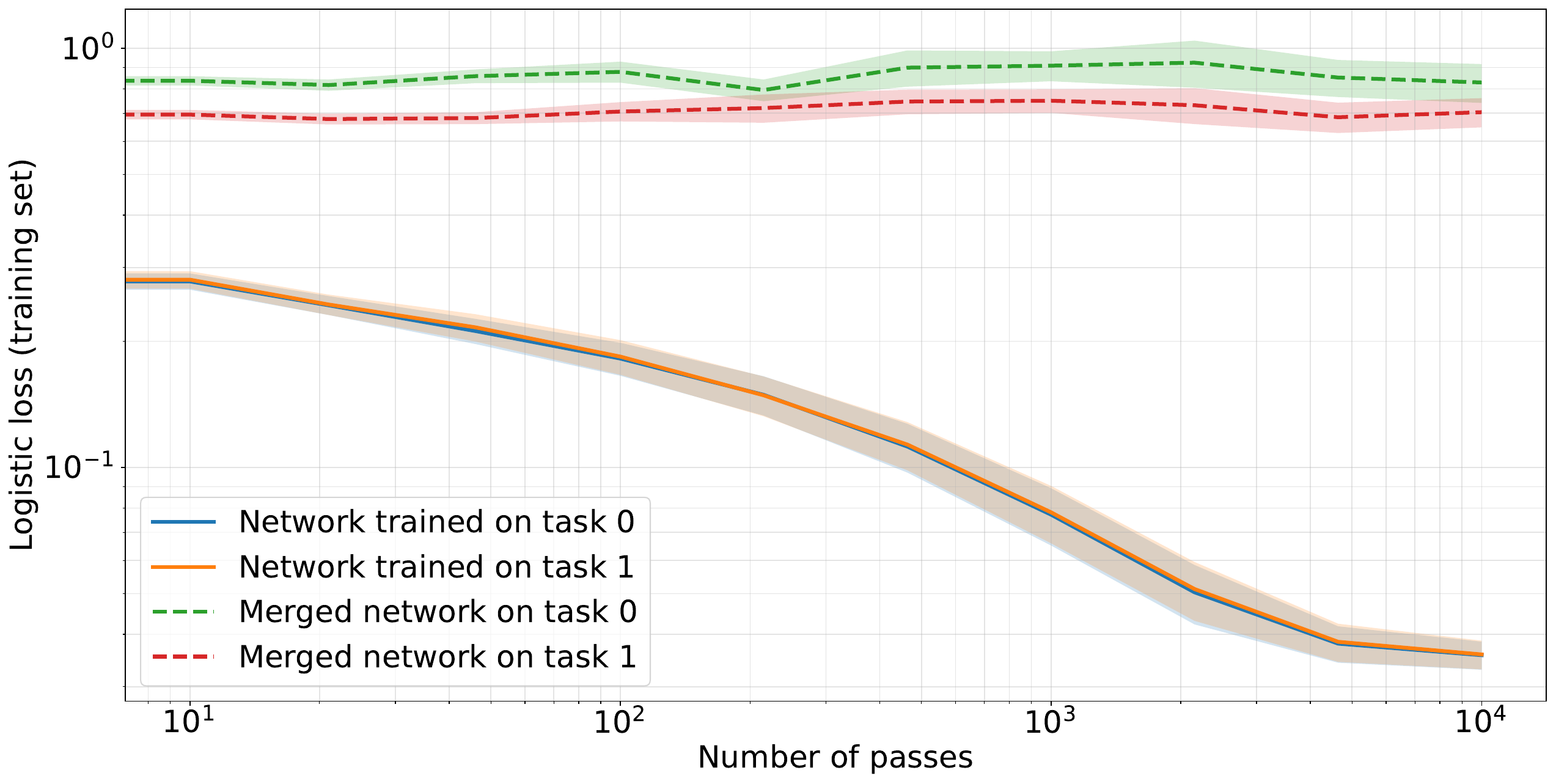}
  \includegraphics[width=0.5\linewidth]{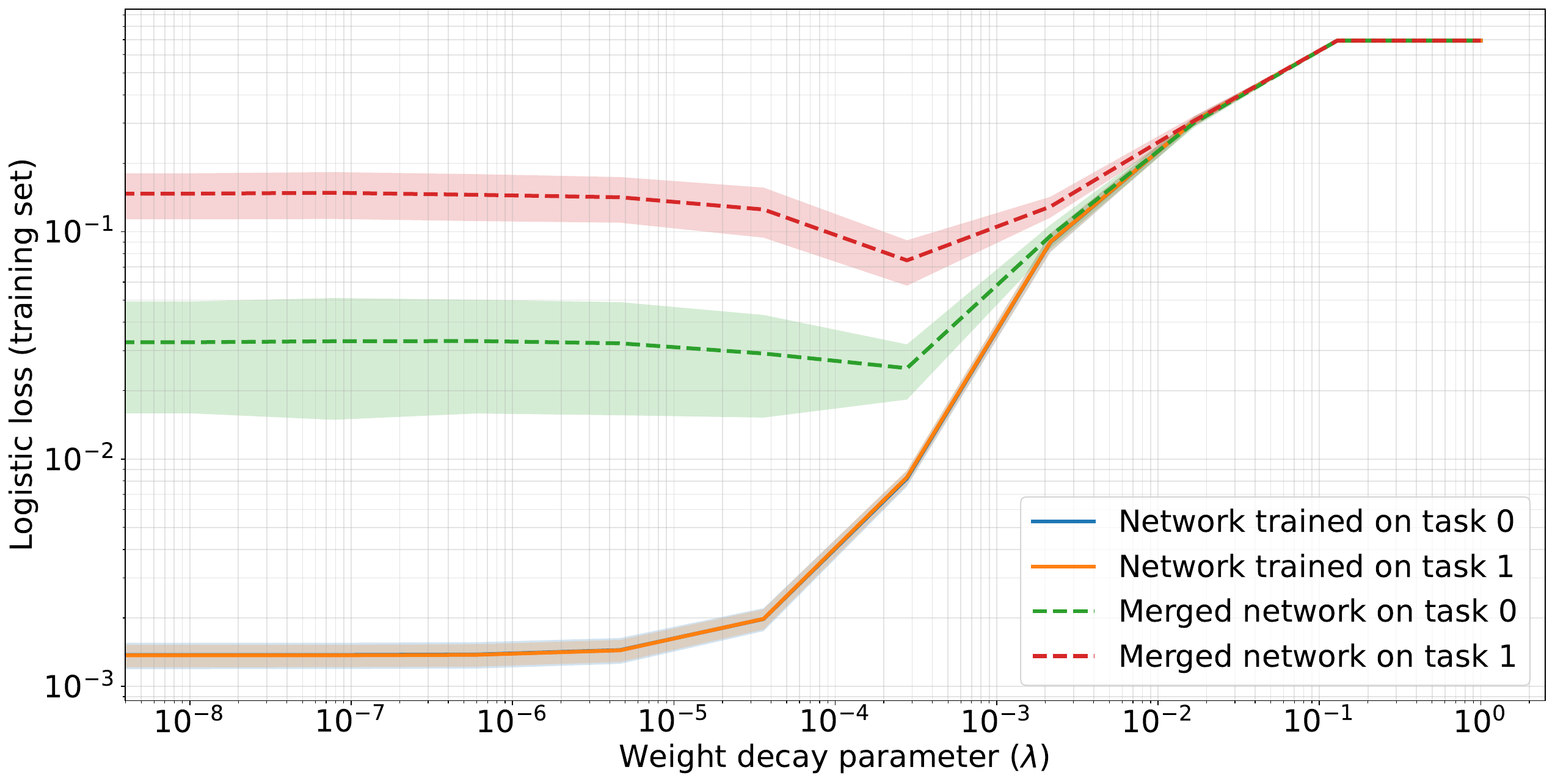}
  \includegraphics[width=0.5\linewidth]{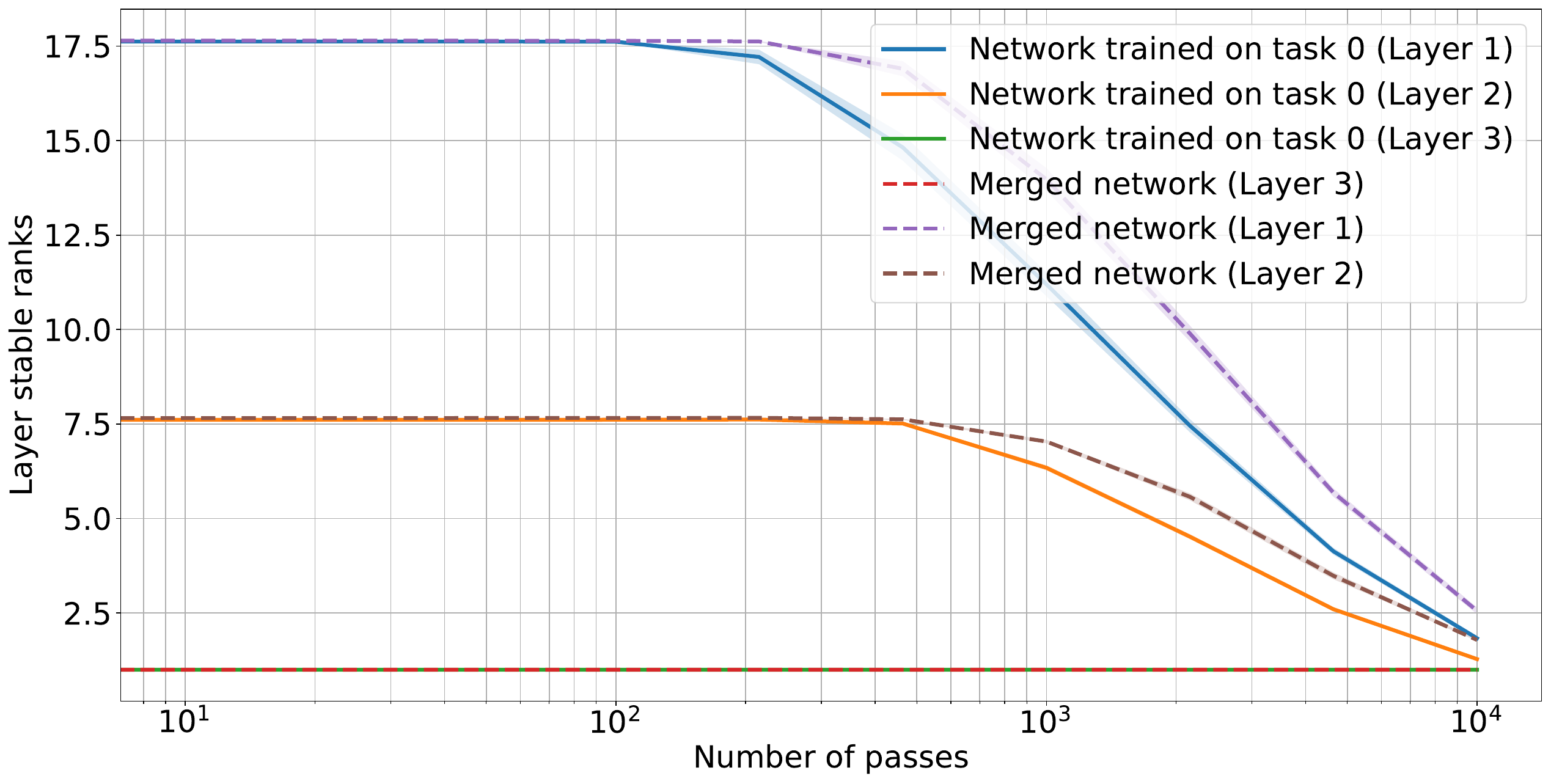}
  \caption{(Fashion MNIST dataset) First row: Training neural networks on different tasks (orthogonal inputs) (left) vs. the same task (right) and
    merging the parameters by adding weight matrices. The resulting network
    performs well on different tasks after sufficiently many iterations, while given the same task, it does not.
    Second row: this effect manifests
    when weight decay strength is sufficiently large (left).
    Stable rank of each weight matrix converges to a small value. Merged
    network matches the stable rank of individual networks (right).}
  \label{fig:t-vs-loss-different-tasks-fm}
\end{figure}

\end{document}